\documentclass[11pt]{article}

\usepackage[final]{acl}

\usepackage{times}
\usepackage{latexsym}

\usepackage[T1]{fontenc}

\usepackage[utf8]{inputenc}

\usepackage{microtype}

\usepackage{inconsolata}

\usepackage{graphicx}
\usepackage{booktabs}
\usepackage{multirow}
\usepackage{subcaption}
\usepackage{siunitx}

\usepackage{tabularx} 
\usepackage{listings}
\usepackage[most]{tcolorbox}
\usepackage{xcolor}

\definecolor{PromptBeige}{RGB}{248,244,230}
\definecolor{PromptConditional}{RGB}{235,245,250} 

\newtcblisting{promptbox}{
    colback=PromptBeige,
    colframe=black!25,
    boxrule=0.5pt,
    arc=3pt,
    left=1.2mm,right=1.2mm,top=1.2mm,bottom=1.2mm,
    breakable,
    enhanced,
    width=\textwidth,
    listing only,
    listing options={
        basicstyle=\ttfamily\footnotesize,
        breaklines=true,
        breakatwhitespace=false,
        columns=fullflexible,
        keepspaces=true,
        showstringspaces=false
    }
}
\newtcblisting{promptboxconditional}{
    colback=PromptConditional,
    colframe=black!25,
    boxrule=0.5pt,
    arc=3pt,
    left=1.2mm,right=1.2mm,top=1.2mm,bottom=1.2mm,
    breakable,
    enhanced,
    width=\textwidth,
    listing only,
    listing options={
        basicstyle=\ttfamily\footnotesize,
        breaklines=true,
        breakatwhitespace=false,
        columns=fullflexible,
        keepspaces=true,
        showstringspaces=false
    }
}

\usepackage{todonotes}

\definecolor{ocr}{HTML}{00C8FF}
\definecolor{ocr}{HTML}{009900}
\definecolor{owenColor}{rgb}{0.0, 1.0, 0.8}
\definecolor{owenurgentColor}{rgb}{0.0, 1.0, 0.8}
\definecolor{jackColor}{rgb}{1.0, 0.6, 0.4}
\definecolor{susanColor}{rgb}{0.5, 0.2, 0.8}
\definecolor{weilingColor}{rgb}{0.2, 0.6, 0.2}
\definecolor{amieColor}{rgb}{0.8, 0.2, 0.3}
\definecolor{peterColor}{rgb}{0.2, 0.2, 0.8}
\definecolor{ForestGreen}{RGB}{20,150,20}

\title{LVLMs and Humans Ground Differently in Referential Communication}

\author{
  Peter Zeng$^{1, 4 }$ \quad
  Weiling Li$^{2}$ \quad
  Amie J. Paige$^{2}$ \quad
  Zhengxiang Wang$^{3,4}$ \\
  {\bf Panagiotis Kaliosis}$^{1}$ \quad
  {\bf Dimitris Samaras}$^{1}$ \quad
  {\bf Gregory Zelinsky}$^{2}$ \quad \\
  {\bf Susan E. Brennan}$^{2}$ \quad
  {\bf Owen Rambow}$^{3,4}$ \\
  $^{1}$Department of Computer Science $^{2}$Department of Psychology\\
  $^{3}$Department of Linguistics $^{4}$Institute for Advanced Computational Science \\
  Stony Brook University \\
    \small{
		\textbf{Correspondence:} \href{pezeng@cs.stonybrook.edu}{pezeng@cs.stonybrook.edu}
	}\\ 
}

\begin{document}
\maketitle

\begin{abstract}
For generative AI agents to partner effectively with human users, the ability to accurately predict human intent is critical. But this ability to collaborate remains limited by a critical deficit: an inability to model \textbf{common ground}. We present a referential communication experiment with a factorial design involving director-matcher pairs (human-human, human-AI, AI-human, and AI-AI) that interact with multiple turns in repeated rounds to match pictures of objects not associated with any obvious lexicalized labels.  We show that LVLMs cannot interactively generate and resolve referring expressions in a way that enables smooth communication, a crucial skill that underlies human language use. We release our corpus of 356 dialogues (89 pairs over 4 rounds each) along with the online pipeline for data collection and the tools for analyzing accuracy, efficiency, and lexical overlap.\footnote{\href{https://github.com/peterzeng/lvlms-referential-game}{https://github.com/peterzeng/lvlms-referential-game}}
\end{abstract}
\section{Introduction}
\label{sec:intro}

Human conversation relies on common ground, accrued and updated by interacting partners \citep{clark1991grounding, clark1986referring, ClarkMarshall1981}. During conversation, partners ground meanings with one another, adapting the referring expressions they use to pick out objects of interest, such that there is less variability \textit{within} a conversation than \textit{between} conversations by different partners discussing the same objects \citep{brennan1996conceptual}.  

\begin{figure}[!t]
    \centering
    \includegraphics[width=\linewidth]{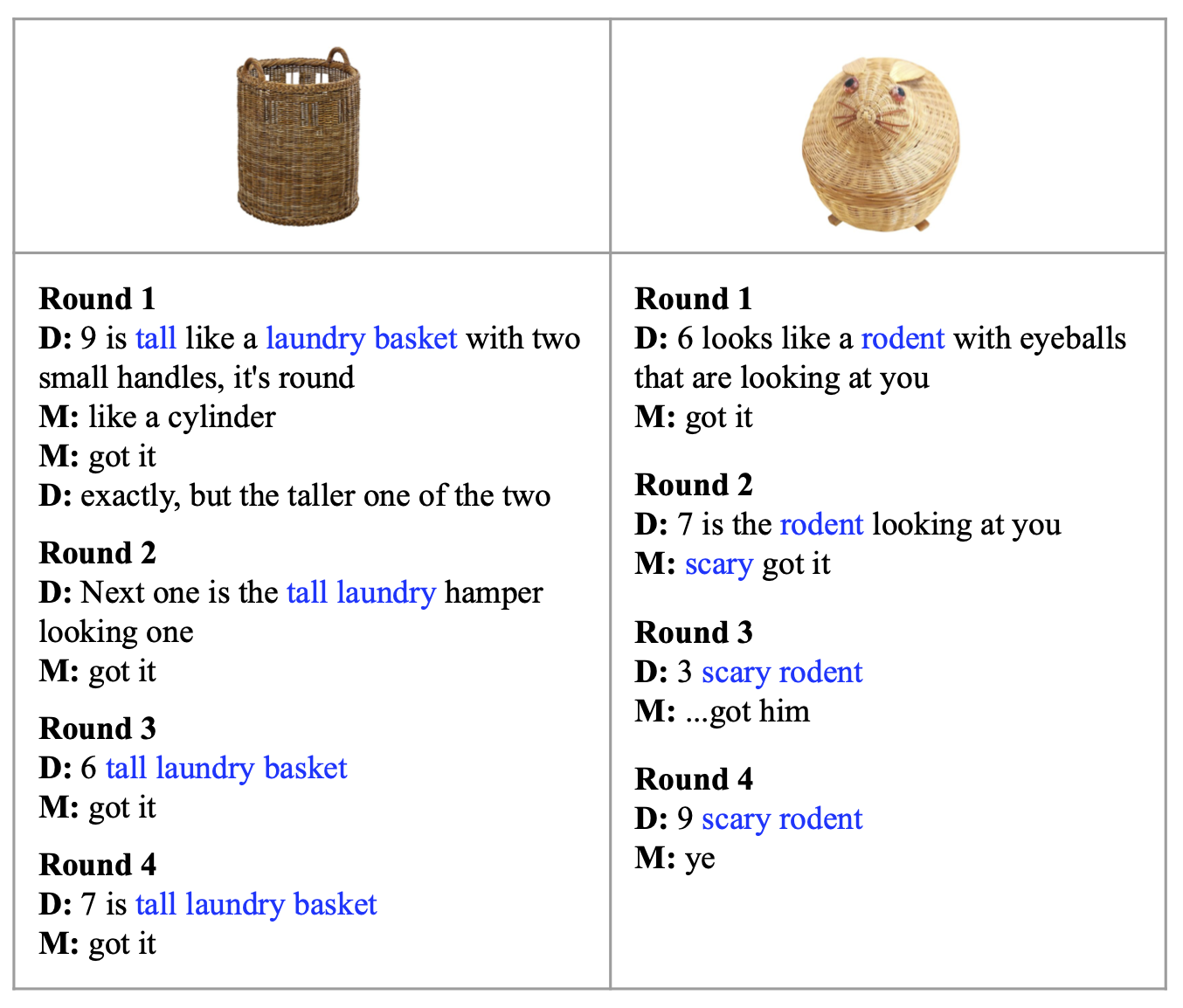}
    \caption{Repeated referring to two baskets (non-lexicalized objects) by a human-human pair, D and M, in Rounds 1-4 of our experiment, with lexical overlap highlighted in blue. Entrainment on more concise language signalling a \textit{conceptual pact} occurs by Round 3, after they consider multiple proposals in Rounds 1-2.}
    \label{fig:H-H examples}
\end{figure}

Conversational partners cannot read one another’s minds, but they can signal that they believe they are talking about the same referent by converging on a referring expression for it (the process of \textit{lexical entrainment}; ibid, see Figure~\ref{fig:H-H examples}). They cannot rely solely on conventional word meanings, as the objects of interest may not be uniquely associated with distinctive labels known to both. They construct meaning jointly by engaging in the process of \textit{grounding}, or seeking and providing evidence about what they mutually believe to be in their common ground. 

Recently, researchers have begun to address the question of whether large language models (LLMs) and large vision language models (LVLMs) engage in grounding, as human discourse partners do \citep{tang-etal-2024-grounding,hua2024talk,imai2025measuring,shaikh-etal-2025-navigating, wang-etal-2025-lvlms}. This question is important both in order to try to understand how models work, and for practical applications in which an AI agent assists a human in a task that requires using language to pick out elements from a visual environment. Studies to date tend to conclude that LVLMs lack the \textit{pragmatic competence} 
needed to coordinate with a partner, and that they struggle in multi-turn conversation.

This paper adds further empirical evidence about the pragmatic competence missing from LVLMs. This is the first study, to our knowledge, to generate (in real time)  multi-turn task-oriented dialogues between pairs of partners in asymmetrical roles (offering different levels of initiative), and that covers all four combinations of human or AI partners filling each role.  We conducted an experiment to examine language use by director-matcher pairs collaborating  to do a referential communication task, based closely on \citet{clark1986referring}’s study that measured how partners accumulate common ground as they match the same objects across multiple rounds. We tested all four combinations of director/matcher roles (human-human, human-AI, AI-human, and AI-AI). This allowed us to quantify not only where LVLMs fail in collaboration, but also to discover why they fail (and in which roles they do so).  
The experimental transcripts are available \href{https://github.com/peterzeng/lvlms-referential-game}{here}.


This paper is organized as follows: Section~\ref{sec:theory} motivates the project within the cognitive science of human communication, and Section~\ref{sec:related} surveys relevant work in human-AI interaction. The experimental design is described in Section~\ref{sec:experiment}, and Section~\ref{sec:results} presents the results along with illustrative examples from the collected dialogues. Section~\ref{sec:follow_up} presents follow-up AI-AI experiments and Section~\ref{sec:conclusion} concludes.

\section{Cognitive Science Background}
\label{sec:theory}

Psycholinguists cannot read the minds of those they study, so they rely on experiments to provide observable evidence about how humans process language. Referential communication is typically studied using matching tasks with carefully chosen stimuli; two partners are provided with a (typically shared) goal such as to discuss and manipulate picture cards, maps, or other visuospatial elements, and their behavior (e.g., words, eye gaze, actions, etc.) is measured. Task roles, available information, and other partner characteristics can be balanced, or else manipulated in order to vary each partner’s initiative or other factors of interest. 

In addition to the process of grounding described in the previous section, other key forces shape dialogues, including pragmatic factors such as \citet{grice1975logic}’s \textit{Cooperative Principle} that governs human expectations when they engage in dialogue, along with its Maxims of \textit{Quality} (“be truthful”), \textit{Quantity} (“say enough but not too much”), \textit{Manner} (“be straightforward”) and \textit{Relation} (“be relevant”). 

Dialogue is also shaped by the costs and affordances of the communication medium 
\cite{clark1991grounding}.
Remarkably, those who share a common purpose (even one given to them by an experimenter) tend to increase their effort as much as it takes for them to accomplish the task, including distributing their individual effort collectively (e.g., depending on their task roles or on the affordances of the communication medium, one partner may work harder to make up for what's difficult for the other). In the context of grounding in different sorts of dialogues, human partners expend the \textit{least collaborative effort} needed to meet a \textit{grounding criterion} sufficient for current purposes \citep{clark1986referring}, which could be perfect accuracy 
in an air traffic control dialogue, high accuracy in an experimental task, or polite engagement while passing the time in a checkout line. 

In an instrumental task, accruing common ground allows human partners to increase their joint efficiency while working together (meaning that they can expend less effort to maintain the same or increasing levels of performance). Figure~\ref{fig:H-H examples} illustrates the emergence of \textit{conceptual pacts}, or temporary, flexible, shared perspectives about referents in conversation \citep{brennan1996conceptual}. These differ from ``conventions'', which exist outside of individual dialogues as word meanings known to the members of a language community. 

Language use by LVLMs, however, is quite a different matter. Training data for LLMs captures statistical regularities that emerge from conventionalized word meanings within texts, rather than from interactively established conceptual pacts.

\section{Related Work in Human-AI Interaction}
\label{sec:related}

Related work on referential communication with LVLMs typically uses human-human corpora as a gold standard for human-AI interaction, finding consistently that humans perform better than models. Although such corpora include spontaneous turn-taking by humans, testing with AI partners tends to not allow multi-turn interaction, so does not test their ability to collaboratively ground meaning or repair misunderstandings.


For example, \textit{PhotoBook}, a widely used dataset for exploring common ground in visually grounded dialogues \citep{haber-etal-2019-photobook}, has human participants play a multi-round online image identification game in which each sees a grid of six visually similar scenes, with some images shared and others visible to only one partner. The partners chat via text and decide which images among a set of highlighted targets are common to both partners or private to only one. \citet{imai2025measuring} adapted PhotoBook to an AI-AI setting using LVLMs and evaluated the AI dyads against the human ones using a suite of proposed grounding metrics. Although LVLM pairs achieved near-human task accuracy, their dialogue differed from human pairs in the formation of common ground, both in efficiency and in lexical adaptation.

\citet{hawkins-etal-2020-continual} had human-human and human-AI director-matcher pairs identify targets within sets that consisted of four photos from the COCO data set, with the goal of being able to use more efficient expressions (dubbed ``ad-hoc conventions'') upon repeated referring to the same photos. Such photos were far more distinctive than the basket targets that we used, making their task much easier (and possibly the labeled COCO data would have been in the models’ training data). \citet{hawkins-etal-2020-continual}’s model adapted over the course of repeated rounds of references to the same objects; however, there was no multi-turn interaction, as human directors simply typed referring expressions and matchers responded (5-10 s later in the case of AI matchers) by selecting the target (before receiving feedback about correctness). 

Using a corpus of nearly 3,000 human-written referring expressions, \citet{tang-etal-2024-grounding} tested model performance of LVLMs as both ``speakers'' and ``listeners'' (analogous to director/matcher roles) in a visual-spatial environment in which the two agents had distinct spatial perspectives. The models performed more poorly at resolving referents (and in both roles) than did humans, but training an open-weight model with evidence of communicative success improved its performance (albeit still not to human-human levels).

\citet{hua2024talk} borrowed \citet{hawkins-etal-2020-continual}’s four-object task and corpus in order to test whether LVLMs could adapt to simulated human partners (as humans do to their real partners) by forming ``ad hoc conventions'' consisting of increasingly efficient referring expressions over rounds of repeated referring. This project used pragmatically-inspired prompting strategies (the best-performing one being to produce shorter and shorter messages with consistent lexical content) rather than any learning on the part of the LVLM. Again, as in \citet{hawkins-etal-2020-continual}, directors presented one-off descriptions, with matcher simply selecting a target, repeated over multiple rounds. LVLMs in the matcher role were often able to interpret the increasingly shortened expressions, but LVLMs in the director role were less able to produce more efficient expressions. For four different LVLM models, accuracy in this simple task ranged from about $40\%$ to ceiling.

\citet{hua2025post} built on this work by developing an interactive (rather than simulated) post-training process to induce ``ad hoc conventions'' (accurate yet concise expressions) through fine-tuning in text-only reference tasks. They used two 
platforms to evaluate their methodology: an interactive form of the game Taboo in which one partner referred (in a single turn) to a target for the other partner (or system) to identify, but without mentioning its conventional name, and a document-based question-answering task that took multiple turns to complete. By evaluating several state-of-the-art proprietary and open-source LLMs, they found that LLMs still lack the ability to spontaneously develop ad-hoc concise referring expressions. Their post-trained LLMs, on the other hand, shortened their messages by up to 26\% in their reference task benchmark and outperformed off-the-shelf counterparts in their document-based task. 

\citet{wang-etal-2025-lvlms} cast LVLMs as overhearers to a transcribed corpus of referential communication produced by humans in both the director and matcher roles, reprising \citet{SCHOBER1989211}'s experiments with human overhearers that found that matchers performed better than overhearers because they could ground meaning interactively whereas overhearers could not. \citet{wang-etal-2025-lvlms}'s LVLM overhearers lagged well behind \citet{SCHOBER1989211}'s human overhearers in accuracy and failed to show any improvement over repeated rounds of the matching task (unlike human overhearers). In another study with text transcripts of human task-oriented dialogue, \citet{sarkar-etal-2025-understanding} showed that LLMs were able to recognize some cues of misalignment (dubbed ``conversational friction”), especially when occurring with clarification questions, but often missed subtler cues. \citet{shaikh-etal-2025-navigating} analyzed grounding in logs of human-LLM interaction (WildChat, MultiWOZ, and Bing Chat) using a taxonomy of ``grounding acts'' \citep{Traum1992AA}. They found that LLMs initiated grounding far less often than humans did, with early failures to ground predicting poor performance later. 

In recent studies using a paradigm similar to \citet{hawkins-etal-2020-continual}'s, \citet{jones2026llms} elicited lexical overlap (that they also referred to as ``convention formation'') within dyads during 50 rounds of referring by directors in which matchers chose single target tangrams out of five successive arrays of 10. After the director's referring expression, the matcher selected a referent (without conversational interaction) and both received feedback about which tangram was selected and whether it matched the intended target. Within human-human, AI-AI, and mixed human-AI dyads, partner roles alternated for each round. 
Results showed that same-type dyads (human-human, AI-AI) improved reliably in accuracy, 
with more lexically consistent expressions 
over rounds, whereas mixed human-AI dyads performed poorly, even when the model was prompted to produce more ``humanlike'' utterances with more concise referring expressions. This suggests that forcing conciseness via prompting is not sufficient to achieve the effects of common ground and conceptual pacts in referring (consistent with \citet{hawkins-etal-2020-continual}'s findings).

\section{Experimental Design and Method} 
\label{sec:experiment}

\subsection{Task Description} 

Our experiment used a referential communication task with human and AI partners paired in director and matcher roles (human-human, human-AI, AI-human, and AI-AI), interacting via multiple, unrestricted text turns. In each of four rounds (with the same director, matcher and set of referents across rounds), the director saw a target sequence of 12 baskets (in a different order for each round) and described them one at a time to the matcher, interacting freely for the matcher to disambiguate each intended referent and move it from a staging area on their screen to the target area to match the director's order. The matcher's set included not only the same 12 baskets from the director's set, but also another 4 so that the last basket(s) would not be trivially easy to identify. This task, based closely on \citet{clark1986referring}, allowed us to observe how partners interactively ground referring expressions and to monitor how they interpret the expressions to resolve or repair meanings and form conceptual pacts across rounds \citep{brennan1996conceptual}. The set of stimulus baskets (Figure~\ref{fig:Basket Sets}) and the interface (Figure~\ref{fig:full_game_ui}) are shown in Appendix \ref{app:prolific}.

To allow direct comparisons across types of partners and roles, all four director-matcher conditions were as similar as possible, hosted online with the platform oTree, an open-source Python package for web-interactive tasks and behavioral research experiments \citep{chen2016otree}. The 3 conditions that involved human partners recruited participants on the platform Prolific \citep{prolific2025}, a high-quality source of vetted and motivated participants who seek to maintain good performance ratings on the platform. As the experiment was online, all communication was via text. The task screen layout had both baskets and chat windows easily visible without scrolling (although scrolling was available to both partners so that they could view prior dialogue from that round). As is typical in interactive human chat media, a series of dots (. . .) was displayed whenever a partner was typing (this too is a useful cue for grounding).
\subsection{Human Participants}
The task was advertised as a job on Prolific that involved communicating with either another human or an AI partner. Given the linguistic demands of the task and need for attention to good task performance for high quality data, we used strict prescreening criteria, restricting participation to fluent English speakers located in the U.S. In addition, to mitigate the quality control challenges inherent in online data collection (e.g., potential cheating or attrition), we updated our filtering criteria to include only participants with a perfect approval record ($100\%$; see Table~\ref{tab:prescreening} for details). The task took an average of one hour to complete (\textit{SD =}28 min). Recruits were told that their partner would be either human or AI, without revealing which.

\subsection{AI Participants\label{sec:ai-participants}}
We selected OpenAI's GPT-5.2 with the ``none'' reasoning option as the LVLM for our AI-AI, human-AI, AI-human experiments \citep{openai2025gpt5}. Following \citet{hua2024talk, wang-etal-2025-lvlms}, we began with GPT-4o when initially developing the prompting. Four rounds of identifying 12 objects turned out to be too complex for that model and required a more capable model. After evaluating several frontier models, we settled on GPT-5.2, first because it was the newest frontier model at the time, and second, for its superior instruction-following capabilities, even at the lowest reasoning level. We used the lowest reasoning level to prioritize responsiveness, as human conversation unfolds under time pressure and humans strive to minimize pauses in conversation \citep{Jefferson1989}.
We designed the system prompts to align with the instructions given to human participants while providing the necessary scaffolding for autonomous interaction. The prompt structure consisted of three key components: (1) Task Context and Role Definition, which mirrors the rules provided to human subjects; (2) Communication Norms, where we explicitly codified pragmatic constraints often implicit to humans (e.g., conciseness, turn-taking, and comparative language) to support natural collaboration; and (3) Structural Scaffolding, which enforced a strict JSON output format to track state updates and induce zero-shot chain-of-thought reasoning \citep{kojima2022large} before generating an utterance or action. In addition, we provided the full history of utterances from both partners, with the matcher receiving an additional image as context that was updated whenever a basket was placed or moved. The full prompts are provided in Appendix \ref{app:prompts}.

\subsection{Resulting Corpus} 


Sessions were removed as invalid when there was evidence of cheating or of not following task instructions; see Table \ref{tab:prescreening} for the proportion removed. A total of 32 human pairs and 39 humans with AI partners successfully completed all four rounds of the task, as well as a post-task survey (totaling 103 human participants that included 58 women, 39 men, and 6 others, with a mean age of $46.6 \text{ years}, \mathit{SD}=13$). Full demographic information is reported in Table \ref{tab:demographics}. The final corpus consists of referential communication dialogues from 32 Human-Human, 22 Human-AI, 17 AI-Human, and 18 AI-AI director-matcher pairs.

\subsection{Evaluation Metrics\label{sec:metrics}}

We evaluated human and AI grounding capabilities using a set of simple metrics that capture the establishment of common ground across three aspects of interaction: communicative success, communication effort,
and lexical entrainment. When reporting results, we first compute each metric for each pair given a specific round and then average the numbers obtained across pairs for that round.


\paragraph{Communicative Success} We measured communicative success using {\bf accuracy}, defined as the percentage of correctly matched baskets within a task round. In matching tasks like this one, humans improve over rounds and reach performance at ceiling, and models perform more poorly, sometimes with no improvement over rounds or declining accuracy.

\begin{figure*}[ht]
    \centering
    \includegraphics[width=1\linewidth]{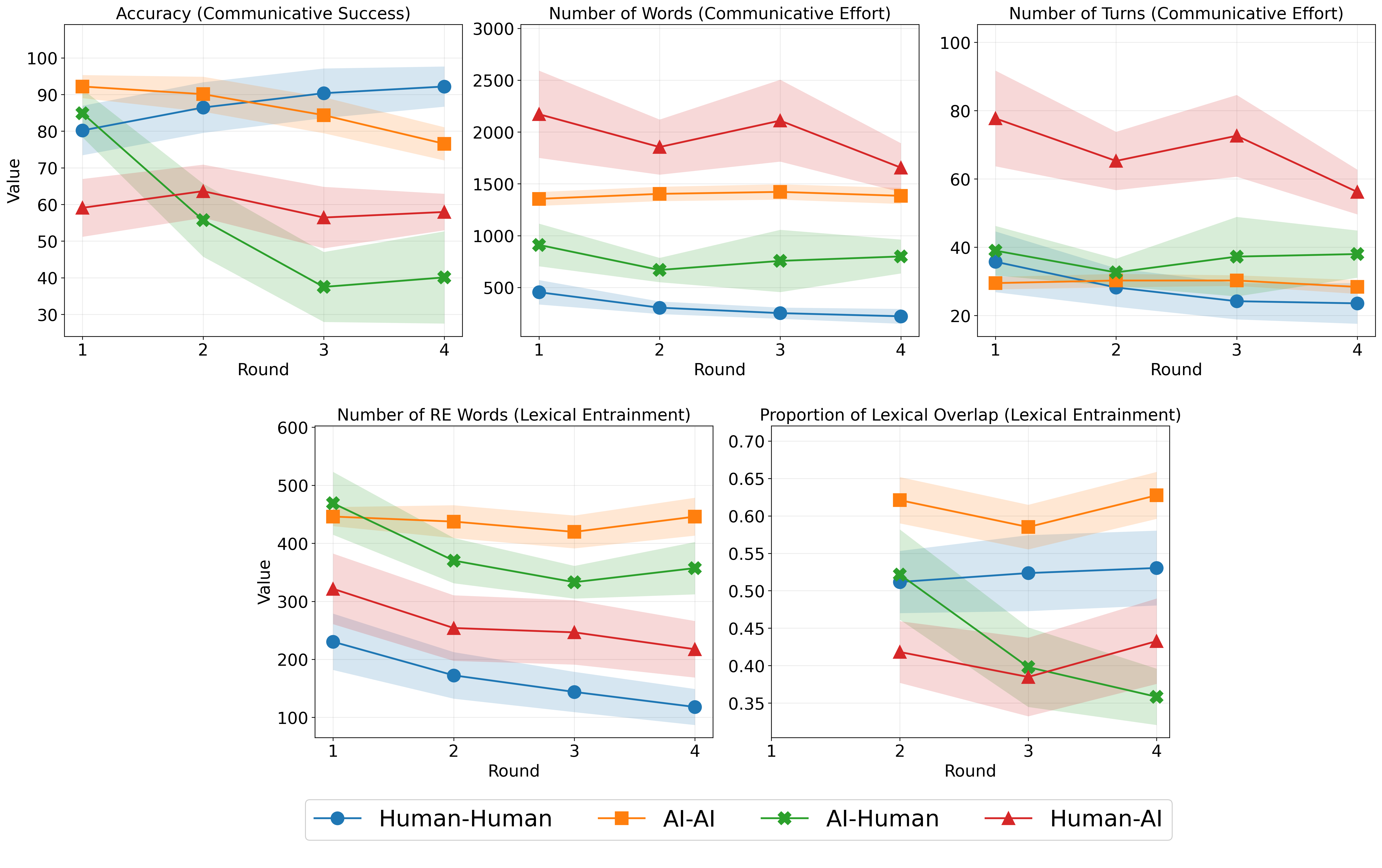}
    \caption{Trends over four rounds for (from left to right) \textbf{accuracy} (\%), \textbf{numbers of words}, \textbf{number of turns}, \textbf{number of words referring expressions}, and \textbf{proportion of lexical overlap with prior rounds} by director--matcher condition. 
    Dots show means with 95\% CIs, with each color denoting a specific pairing condition.}
    \label{fig:ols_accuracy_efficiency_entrainment}
\end{figure*}

\paragraph{Communicative Effort}
As conversational partners progressively establish common ground, they require less effort to communicate. We quantified this using the {\bf number of words} and the {\bf number of turns} in each task dialogue as proxies, capturing overall communication cost and back-and-forth coordination, respectively. Other related measures (e.g., number of utterances) patterned similarly, 
and are reported in Appendix~\ref{app:more_effort_metrics}.

\paragraph{Lexical Entrainment} Lexical entrainment refers to the tendency of interlocutors to reuse lexical material (in human conversation, this happens in a more concise manner) when referring to previously established objects \citep{brennan1996conceptual, garrod1987}. We operationalized this concept using two metrics that capture length reduction and lexical overlap of referring expressions across rounds. 

Let $\mathrm{RE}_{i}^{(b)}$ denote the referring expression or description for basket $b$ in round $i$, and let $\mathrm{Tok}(\mathrm{RE}_{i}^{(b)})$ denote its sequence of content words. We defined the length of a referring expression as the number of content words, i.e., as  $|\mathrm{Tok}(\mathrm{RE}_{i}^{(b)})|$ ({\bf number of RE words}).  

To quantify lexical overlap, we computed round-$i$ relative lexical overlap (RLO$_i$, {\bf proportion of lexical overlap}) as the multiset overlap of content words ($\mathrm{Inters}(A, B)$) between the referring expression used in round $i$ and the one used in round $i-1$. RLO always ranges between 0 and 1 and is defined as follows.

\[
\mathrm{RLO}_i^{(b)} =
\frac{|\mathrm{Inters}(\mathrm{Tok}(\mathrm{RE}_{i-1}^{(b)}), \mathrm{Tok}(\mathrm{RE}_{i}^{(b)}))|}
{|\mathrm{Tok}(\mathrm{RE}_{i}^{(b)})|}
\]
This lexical overlap measure is similar to other measures used in the literature; we discuss these in Appendix~\ref{app:more_lexical_entrainment_metrics}.  We use our definition because it ignores word order and has an immediate intuitive interpretation aligned with entrainment.  While higher overlap reflects greater lexical reuse, it does not by itself imply increased efficiency; we therefore interpret overlap jointly with length-reduction trends when assessing lexical entrainment.

\paragraph{Referring Expression Extraction}
We used GPT-5 \citep{openai2025gpt5} to automatically extract referring expressions (REs) for each target basket at the dialogue level. We validated the extracted expressions against human annotations on a comparable corpus, obtaining an F$_1$ score of 0.86 under ROUGE-L \citep{lin-2004-rouge}. See Appendix~\ref{app:re_extraction_validation} for details of the validation setup and results.



\begin{table}[!h]
\centering
\small
\definecolor{posColor}{RGB}{26,127,55}
\definecolor{negColor}{RGB}{209,36,47}
\definecolor{zeroColor}{RGB}{110,119,129}
\definecolor{naColor}{RGB}{110,119,129}
\begin{tabular}{lcccc}
\toprule
 & HH & AA & AH & HA \\
\midrule
Accuracy& \textcolor{posColor}{4.0**} & \textcolor{negColor}{-5.3***} & \textcolor{negColor}{-15.3***} & \textcolor{negColor}{-1.1} \\
\# Words  & \textcolor{negColor}{-74.9***} & \textcolor{posColor}{10.6} & \textcolor{negColor}{-24.7} & \textcolor{negColor}{-129.1} \\
\# Turns & \textcolor{negColor}{-4.1**} & \textcolor{negColor}{-0.3} & \textcolor{posColor}{0.2} & \textcolor{negColor}{-5.7*} \\
\# RE Words  & \textcolor{negColor}{-36.6***} & \textcolor{negColor}{-1.8} & \textcolor{negColor}{-37.3***} & \textcolor{negColor}{-32.0*} \\
L Overlap & 0.0 & 0.0 & \textcolor{negColor}{-0.1***} & 0.0 \\
\bottomrule
\end{tabular}
\caption{Ordinary least squares (OLS) regression slopes ($c$) of each metric across rounds by director-matcher condition. H: human. A: AI. Green \textcolor{posColor}{$c>0$}, red \textcolor{negColor}{$c<0$}; stars: *** $p<0.001$, ** $p<0.01$, * $p<0.05$.}
\label{tab:ols_metrics_trends}
\end{table}

\section{Results\label{sec:results}}
Figure~\ref{fig:ols_accuracy_efficiency_entrainment} 
shows round-by-round trends across director–matcher conditions with means and 95\% confidence intervals for five evaluation metrics that capture the three aspects of interaction relevant to grounding capabilities (accuracy, effort, and lexical entrainment). We used ordinary least squares (OLS) regression to quantify the overall trends for these metrics and report the slopes along with the respective significance levels in Table \ref{tab:ols_metrics_trends}.

\paragraph{Human-Human}

Human-human pairs achieved high communicative success (accuracy), starting at 80\% and increasing steadily over rounds to over 90\% by Round 4. 
In terms of communicative effort, the numbers of words and turns per round started relatively low and decreased over rounds.  For the lexical entrainment measures, the number of RE words used to describe each basket was also low and decreased, and the REs consistently re-used a high percentage of lexical material from the previous round (above 50\%).  

For example, the pair who contributed the dialogues in Figure~\ref{fig:H-H examples} completed all four rounds of the task in only 28 min and with only two errors (during the first round). Both dialogues show evidence of entrainment by the third round (with the matcher contributing lexical content to the resulting ``scary rodent'' conceptual pact). Unlike the AI partners, human partners sometimes explicitly acknowledged their common ground with meta-linguistic references, as another pair did with descriptions like ``the one we worked hard on last time'' (See Figure \ref{fig:H-H-example2}).

The quantitative data and corpus show clearly that humans entrained: the REs they used got shorter, but remained related to the previous round's RE (lexical overlap) and overall, the interactions became 
shorter (in both words and turns) and more efficient over rounds, meaning that the partners expended less linguistic effort to understand each other increasingly well.  These results are consistent with previous referential communication experiments \citep[e.g.,][]{brennan1996conceptual, clark1986referring}, despite our novel setup (online and typing-only).

\paragraph{AI-AI} The AI-AI interactions showed a very different pattern from the human-human interactions.  Communicative success (accuracy) started high, at 90\% (even higher than for human-human interactions), but decreased 
steadily over rounds.
As for communicative effort, the number of words used began at about three times that of human-human pairs in Round 1 and did not decrease; turns, as well, were much longer in this condition than in the human-human condition. 
That the number of turns did not decrease over rounds suggests that the AI-AI pairs failed to benefit from common ground and that they continued to engage in excessive confirmation, even in later rounds (for an example, see Figure~\ref{fig:AI-AI-example}).
Finally, concerning lexical entrainment, the number of RE words started out high (around 450) and did not decrease. Lexical overlap remained high as well (always at least 60\%).  

The transcripts show that AI directors were strikingly verbose, presenting descriptions in long turns rather than in the incremental installments that humans typically use. AI matchers typically accepted AI directors' presentations by  repeating them in their entirety, e.g., \textit{``Placed the tall open cylindrical hamper-style basket with two small loop handles near the rim into slot 9.”}) in contrast to human matchers, who almost always responded with a brief confirmation (e.g., “done”, “ok”, or “got it” as in Figure \ref{fig:H-H examples}). Our prompting strategy (see Appendix~\ref{app:prompts}) included instructions about pragmatics such as a suggestion to confirm when necessary, but AI partners did this routinely and excessively regardless of need; some sessions included confirmation questions at the end of each and every turn.  (See \citep{brennan1995interaction}
for discussion of how modeling a \textit{grounding criterion} \citep{clark1986referring} could avoid this.) Sometimes AI directors appeared to respond appropriately to matchers’ routine confirmation questions and a repair ensued, but then the AI matcher still selected the wrong basket. And sometimes in subsequent re-referring using the same expression that succeeded earlier, the wrong basket was placed. Future work will involve systematic analyses of repair on this corpus.

Not infrequently, the AI director would inexplicably add entirely new modifying words or phrases to a description (even when the match had been successful in the prior rounds) or simply seem to switch to describing the wrong basket (a hallucination, perhaps due to interference from other baskets or losing track of which basket was the current target). Strangely, sometimes the AI matcher placed the target correctly even though the director’s description was incorrect (see Figure~\ref{fig:AI-AI-example} for an example).
We note also that occasionally, in later rounds, the LVLM would begin describing baskets in the wrong order. This contributed to the drop in accuracy across rounds.

We conclude from these results that the AI pairs did {\em not} entrain on conceptual pacts: the REs didn't become shorter, overall effort failed to decrease, and accuracy did not improve (in fact, it decreased significantly). The high lexical overlap in this condition does not contradict this interpretation, as the AI director's very long REs did not vary much across rounds. Overall, the AI participants did not use their communication history to track what information their partner needed, nor adapt their communication strategy to increase efficiency (violating \citet{grice1975logic}'s Maxim of Quantity).

\paragraph{Human-AI}  
In these mixed pairs, the human was the director. This condition showed the least accuracy at Round 1 out of all four conditions, and accuracy did not increase over rounds. At the same time, the communicative effort was much higher than for the other conditions, for both words and number of turns, across all four rounds (and with effort decreasing only slightly by Round 4). Concerning lexical entrainment, the human director did not use many RE words to describe the baskets, averaging only about 100 more than a human director with a human matcher, and this decreased over rounds.  However, lexical overlap was low.

Figure~\ref{fig:H-AI-example} shows dialogue from a human-AI pair that illustrates what appears to be the human director's flexible attempts to ground, alongside the AI matcher's  persistence in re-introducing terms from previous rounds. Given this basket's distinctiveness within the set (see Figure~\ref{fig:Basket Sets}), the excessive confirmations (underlined) violate \citet{grice1975logic}'s Maxim of Quantity.


\paragraph{AI-Human} In this condition, the director was the LVLM, and the matcher, human. Accuracy started high (85\%) but decreased precipitously, unlike in any of the other conditions. Communicative effort started out \textit{much} lower than in the other mixed condition (with an AI director) but with twice as many words as in the human-human condition and roughly the same number of turns. However, unlike with human pairs, effort did not decrease over rounds. As for the lexical entrainment measures, the number of RE words started high, as in the AI-AI condition, but decreased in Round 2 and then leveled off. However, lexical overlap with the previous round, while starting in Round 2 at same level as for human-human pairs, dropped rapidly in Rounds 3 and 4, meaning that more and more new lexical material was (needlessly) introduced within the REs during the later rounds. This is evidence not only of a lack of entrainment, but coupled with the abysmal accuracy in this condition, a lack of successful communication.

The causes of this accuracy decrease appear to be twofold: we observed that after Round 1, the AI director would occasionally begin describing the baskets in the wrong order. This, in addition to the lack of lexical entrainment, contributed to the large decrease in accuracy across rounds. 

Overall, human matchers (whether in mixed pairs or in human-human pairs) tended to respond with concise acknowledgments (e.g., ``got it'' just before placing a basket). 
However, in cases where human matchers initially tried asking clarifying questions of their AI directors, they sometimes appeared to realize that their partner was not a capable communicator and gave up. A sample AI-human dialogue in which the human matcher tried unsuccessfully to repair can be found in Figure \ref{fig:AI-H-example}.

\paragraph{Participants' Perceptions and Experiences} 

Human participants' post-task survey responses were informative about their subjective experiences of the task and of their partners (human or AI). These self-reported measures, summarized in Table \ref{tab:Survey data}, align with (and may help explain) the objective performance patterns we observed.

Consistent with the performance gap, human partners were rated significantly higher than AI partners across all dimensions related to collaboration (all \textit{p}s < .001). In human-human pairs, ratings for capability, helpfulness, and adaptability approached ceiling (4.25-4.88 on a 5-point scale).  
In contrast, ratings for AI partners were consistently
lower (\textit{M} < 3.0; \textit{p}s < .001). Participants in the AI-human condition reported the lowest score for ``Collaboration Improvement'' (\textit{M} = 2.12), reinforcing the finding that human matchers struggled to establish common ground (conceptual pacts about referents) with AI directors. Participants in the human-human condition also perceived their partners as significantly more human-like (\textit{M} = 77.25\%) than those in the mixed conditions (i.e., human-AI and AI-human; \textit{M} = 12.58\%; \textit{p} < .001).

There were no significant differences between the human-human and mixed conditions in prior AI familiarity (\textit{M} = 3.77 vs. 3.41) or usage frequency (\textit{M} = 3.94 vs. 3.56; \textit{p} > .05), indicating that prior exposure to AI did not account for observed effects.

\paragraph{Summary}  These results are consistent with previous findings in the referential communication literature and with the theory laid out in Section~\ref{sec:theory}: that humans propose expressions incrementally for their partners to resolve and ratify while decreasing communicative effort and increasing efficiency. AI discourse participants can follow prompts to confirm and repair, but do not adapt this behavior to the needs of a partner.  For the AI-AI condition, this was not as problematic as for the mixed pairs,
as presumably neither AI partner expected grounding to happen (or become frustrated when it didn't).

In mixed pairs, the results differed depending on who was in the director role (where more initiative and linguistic effort is needed). When AI takes most of the initiative, it is often inaccurate; when a human partner takes the initiative, enormous effort may be required. The post-task survey of human participants demonstrated their significant frustration when paired with an AI partner as opposed to a human partner.  LVLMs' inability to flexibly adapt to human partners is 
a concerning finding indeed, as major applications of dialogue technology involve AI systems supporting humans (rather than AI systems talking to each other).

\section{Follow-Up AI–AI Experiments\label{sec:follow_up}}

\begin{table}[t]
\centering
\scriptsize
\setlength{\tabcolsep}{3.0pt}
\definecolor{posColor}{RGB}{26,127,55}
\definecolor{negColor}{RGB}{209,36,47}
\definecolor{zeroColor}{RGB}{110,119,129}
\definecolor{naColor}{RGB}{110,119,129}
\begin{tabular}{lcccc}
\toprule
Round Index $\rightarrow$ & 1 & 2 & 3 & 4 \\
Condition $\downarrow$ &  &  &  &  \\
\midrule
Default & 92.2 & 90.1 {\tiny\textcolor{negColor}{(-2.1)}} & 84.4 {\tiny\textcolor{negColor}{(-5.7)}} & 76.6 {\tiny\textcolor{negColor}{(-7.8)}} \\ \midrule \midrule

Simple Prompt & 100.0 & 83.3 {\tiny\textcolor{negColor}{(-16.7)}} & 83.3 {\tiny (0.0)} & 75.0 {\tiny\textcolor{negColor}{(-8.3)}} \\ \midrule \midrule

Low Reasoning & 100.0 & 91.7 {\tiny\textcolor{negColor}{(-8.3)}} & 100.0 {\tiny\textcolor{posColor}{(+8.3)}} & 91.7 {\tiny\textcolor{negColor}{(-8.3)}} \\
Medium Reasoning & 91.7 & 91.7 {\tiny (0.0)} & 100.0 {\tiny\textcolor{posColor}{(+8.3)}} & 91.7 {\tiny\textcolor{negColor}{(-8.3)}} \\
High Reasoning & 91.7 & 91.7 {\tiny (0.0)} & 83.3 {\tiny\textcolor{negColor}{(-8.4)}} & 100.0 {\tiny\textcolor{posColor}{(+16.7)}} \\ \midrule \midrule

GPT-5.2 vs Gemini Pro & 91.7 & 70.8 {\tiny\textcolor{negColor}{(-20.9)}} & 75.0 {\tiny\textcolor{posColor}{(+4.2)}} & 50.0 {\tiny\textcolor{negColor}{(-25.0)}} \\
Gemini Pro vs GPT-5.2 & 83.3 & 83.3 {\tiny (0.0)} & 75.0 {\tiny\textcolor{negColor}{(-8.3)}} & 50.0 {\tiny\textcolor{negColor}{(-25.0)}} \\
GPT-5.2 vs Claude & 25.0 & 41.7 {\tiny\textcolor{posColor}{(+16.7)}} & 50.0 {\tiny\textcolor{posColor}{(+8.3)}} & 8.3 {\tiny\textcolor{negColor}{(-41.7)}} \\
Claude vs GPT-5.2 & 33.3 & 50.0 {\tiny\textcolor{posColor}{(+16.7)}} & 50.0 {\tiny (0.0)} & 62.5 {\tiny\textcolor{posColor}{(+12.5)}} \\
\bottomrule
\end{tabular}
\caption{Accuracy across rounds in the follow-up AI-AI experiments. \textcolor{posColor}{Green} and \textcolor{negColor}{red} indicate positive and negative changes in accuracy from the previous round to the current round, respectively. The ``Default'' condition corresponds to the main experiments; unless otherwise specified, both the director and the matcher are GPT-5.2 with reasoning effort set to ``none".}
\label{tab:more_ai_experiments_accuracy}
\end{table}

Our main experiments considered only AI-AI pairs drawn from the same model, using a fixed prompt and a fixed reasoning effort. To assess the robustness of our main findings, we conducted follow-up experiments under the three conditions, as discussed below. Guided by the results of the main experiments, which found minimal variation in the AI-AI results, as well as for budget reasons, we ran each condition only up to two times. 


Table~\ref{tab:more_ai_experiments_accuracy} reports accuracy across rounds in the additional experiments. Consistent with the main experiments, AI-AI communication remained verbose and exhibited repeated referring expressions across rounds (see Tables~\ref{tab:more_ai_experiments_communication_effort} and~\ref{tab:more_ai_experiments_lexical_entrainment} in Appendix~\ref{app:follwup_results}). We therefore focused our analysis here on accuracy.

\paragraph{A Simplified Prompt} We first removed the elaborate communication norms and zero-shot CoT instruction from the default prompt (Section~\ref{sec:ai-participants}). Consistent with the main experiments, AI-AI accuracy still declined across rounds (100.0 $\rightarrow$ 75.0; Table~\ref{tab:more_ai_experiments_accuracy}), showing no evidence of grounding.

\paragraph{Varying Reasoning Efforts} Next, we varied GPT-5.2’s reasoning effort from low to high. Although higher effort sometimes yields transient recoveries (e.g., 100.0 in Round 3 for low/medium; 100.0 in Round 4 for high), accuracy did not improve monotonically across rounds (Table~\ref{tab:more_ai_experiments_accuracy}), showing no evidence of common-ground formation.

\paragraph{Mixed AI Pairs} Last, we paired GPT-5.2 with Gemini-3-Pro \citep{google2025gemini3} (minimal reasoning) and Claude Opus-4.5 \citep{anthropic2025opus45} (low reasoning), swapping director and matcher roles. Mixed-model pairs consistently underperformed same-model pairs and showed no steady gains in accuracy across rounds. The only partial exception was Opus-4.5 versus GPT-5.2, where accuracy increased overall, but remained well below human-human performance by Rounds 3 and 4.
Dialogues between Opus-4.5 and GPT-5.2 pairs were also substantially more verbose than other pairing (see Table~\ref{tab:more_ai_experiments_communication_effort} in Appendix~\ref{app:follwup_results}), suggesting the limited 
utility of verbosity in effective coordination.

\section{Conclusion}
\label{sec:conclusion}


We have described a carefully designed factorial experiment in which human and/or AI partners cooperated on a referential communication task. We summarize the main insights here.

\textbf{Having an AI partner in either (or both) of the director-matcher roles made a pair less accurate \textit{and} less efficient.}
Although pairs with AI directors started out as equally accurate (or even numerically more accurate) than human pairs in Round 1, accuracy declined precipitously for AI-human pairs and gradually for AI-AI pairs. Despite poor accuracy, pairs with AI directors were remarkably verbose, producing many more words than did pairs with human directors. Only human-human pairs showed significant improvement in accuracy over rounds, accompanied by increased efficiency over rounds. These trends are consistent with the ability of human-human pairs to rapidly establish common ground and entrain on compact, reusable referring expressions that reflect conceptual pacts.

\textbf{In contrast to humans, even a frontier LVLM, GPT-5.2, showed no hint of any ability to build common ground.}
This was true even when the partner was another instance of the same model. AI-AI accuracy decreased significantly across rounds, while effort, as measured by words and turns, remained largely flat. 
Overall, LVLMs did not appear to track or exploit common ground, regardless of role or partner type, and despite having access to the dialogue history.
Accuracy collapsed for the AI-human pairs in which the AI partner performed the director role, suggesting that there may be substantial risks for embodied AI when it is expected to take initiative in a collaborative or human-facing task.

\textbf{In an interactive collaborative task, humans can recognize whether their partner in a referential communication task is AI or not}. If this recognition is (at least partially) because of the failure of the AI partner to ground, then the failure to accrue common ground makes GPT-5.2 fail the Turing Test.


\section*{Limitations}

This study was conducted only in English, with only one type of object (not associated with conventionally lexicalized labels), and with only one LVLM for the full factorial design (GPT-5.2). Only proprietary SOTA models were used for limited followup studies, as those tend to perform better than open models that have not been fine-tuned or trained (which is beyond the scope of this study). That open-weight models were not considered may affect reproducibility of results.

Maintaining the quality of Prolific data collection requires monitoring in order to respond to occasional queries from participants, as well as some judgment calls about task completion and payment (e.g., participants who cheated were not paid, but their (human) partners were). We report our success and strategies for recruitment in Table~\ref{tab:prescreening}.

Our analytic method for measuring lexical entrainment is a proxy for the kinds of painstaking coding done by human coders, likely underestimates entrainment, as it does not capture the full range of decisions that human coders make. The means shown in Figure~\ref{fig:ols_accuracy_efficiency_entrainment} do not identify the precise turn in which a particular pair has achieved a conceptual pact about a particular object. We have not yet analyzed all the transcripts for conversation repair or for so-called ``hallucinations.'' We invite readers with an interest in additional analyses to examine the corpus, which is publicly available at \href{https://github.com/peterzeng/lvlms-referential-game}{https://github.com/peterzeng/lvlms-referential-game}.

Regarding our prompting, we recognize that such explicit interventions are highly setup-specific and may not generalize beyond this experimental task. While this scaffolding was necessary to enable the LVLMs to successfully flow through and submit the task, such interventions can lead to unintended model behaviors. Ultimately, we concur with recent findings \citep{hua2024talk} that prompt engineering is not likely to be the solution to better human-AI collaboration.  

\section*{Ethics Statement}

\textbf{Human Subjects} The corpus described here was
collected with approval from 
our institution's committee on research involving human subjects
and with informed consent provided by the participants. The corpus contains no personally identifiable information.  The Prolific workers were paid \$12 per hour.


\section*{Author Contributions}
Peter Zeng co-managed the experimental design, developed the experimental interface as well as managed the backend, led the data collection, finalized prompts, and contributed to writing the Introduction, Related Work, Experimental Design, and respective sections in the Appendix.

Weiling Li and Amie Paige co-managed the experimental design and data collection, designed cognitively based prompts, performed statistical and qualitative analyses, and contributed to writing the Cognitive Science Background, Methods, and Results sections. Amie Paige created and curated the stimuli for the referential communication task, as well as suggested the validation corpus for referring expression extraction.

Zhengxiang Wang analyzed and visualized the quantitative results from the experiments and drafted the following sections: Metrics, Follow-Up AI-AI Experiments, Conclusion, and the related parts in the Appendix. 

Panagiotis Kaliosis tested the interface for the two-player collaborative  game, handled and drafted the Related Work in Human-AI Interaction section. 

Susan Brennan and Owen Rambow were the primary supervisors on this paper. They provided feedback on the experimental design, task interface, and prompting methods, conducted extensive testing of the interface, and edited the the paper to ensure consistency in theory and terminology. Susan Brennan wrote the Cognitive Science Background section and parts of the Introduction, Experimental Design and Method, and Related Work sections. Owen Rambow developed and iterated on the metrics, contributed to Results, and wrote the Introduction and Conclusion sections. 

Dimitris Samaras and Gregory Zelinski contributed to the conceptualization of the experiments and the analyses.

\section*{Acknowledgements}
This material is based upon work supported by the National Science Foundation under Grant No. 2125295    and by a seed grant from Stony Brook University. Any opinions, findings, and conclusions or recommendations expressed in this material are those of the author(s) and do not necessarily reflect the views of the National Science Foundation.

Peter Zeng, Zhengxiang Wang, and Owen Rambow are grateful for support from the Institute for Advanced Computational Science (IACS) at Stony Brook University, in particular the free GPT access it provides.

\bibliography{custom}

\appendix

\section{Prolific Data Collection} 
\label{app:prolific}

\noindent\textbf{Experimental Flow:}
\begin{enumerate}
    \item Participants review instructions/complete consent form
    \item Participants are matched with a partner
    \item Participants complete 4 rounds of the task with the same partner, and their roles remain fixed throughout. The order of the target baskets varies across rounds.
    \begin{enumerate}
        \item Partners communicate via chat
        \item NOTE: the Matcher submits the ordered baskets each round
    \end{enumerate}
    \textit{Inter-round}: participants review feedback and complete attention checks
    \item After the final round, participants respond to questions about...
    \begin{enumerate}
        \item how well their partner collaborated with them (Likert and free response)
        \item whether they believed their partner was AI (scale and free response)
        \item their personal AI use (multiple choice)
    \end{enumerate}
    \item Debriefing form/Return to Prolific link
\end{enumerate}

\begin{figure*}
    \centering
    \includegraphics[width=\linewidth]{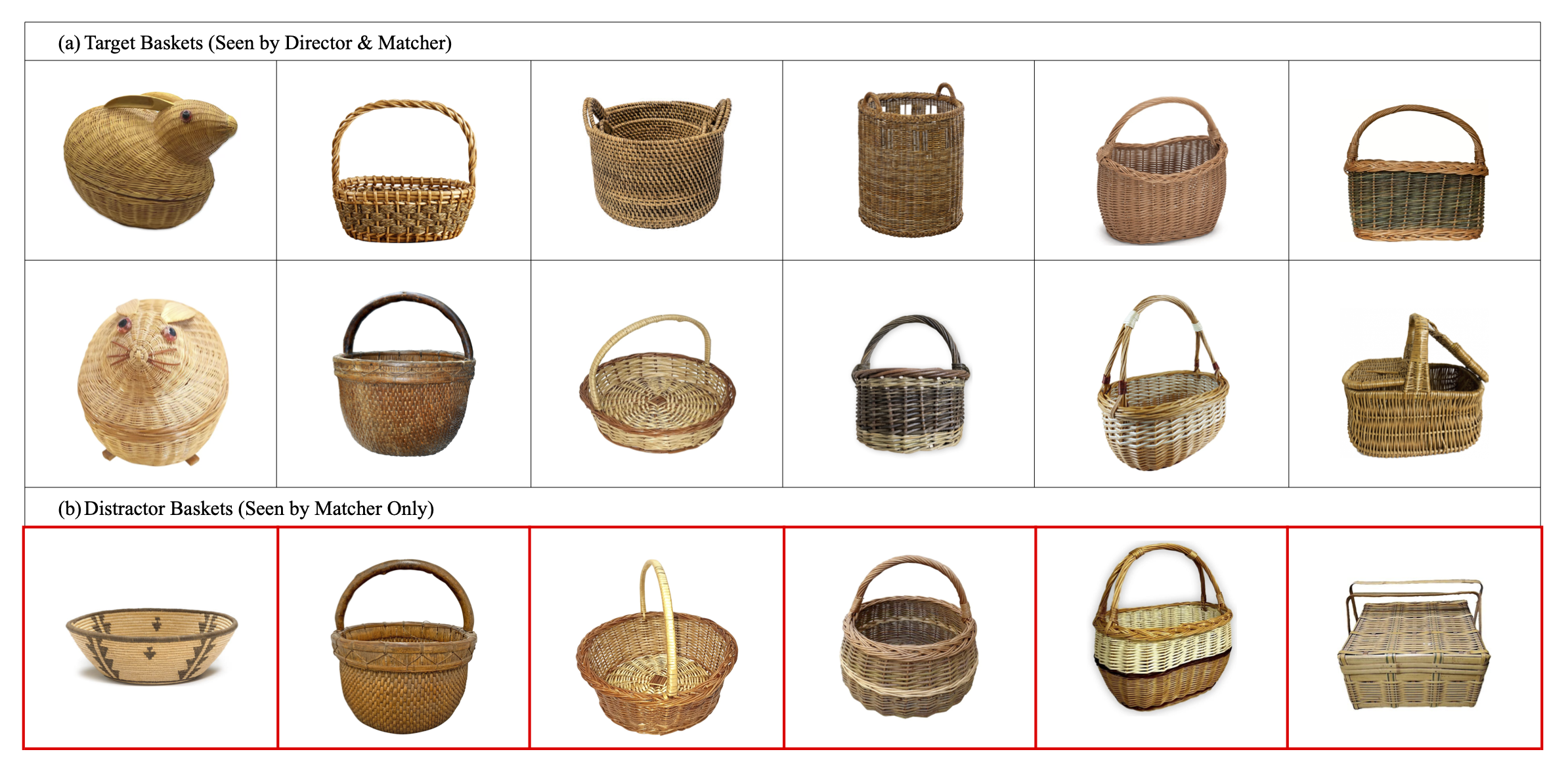}
    \caption{Complete stimulus set used in the task. (a) The 12 target baskets viewed by both the director and the matcher. (b) The 6 distractor baskets viewed only by the matcher, mixed with the targets.}
    \label{fig:Basket Sets}
\end{figure*}

\begin{figure*}[t!]
    \centering
    \begin{subfigure}[t]{0.48\textwidth}
        \centering
        \includegraphics[width=\linewidth]{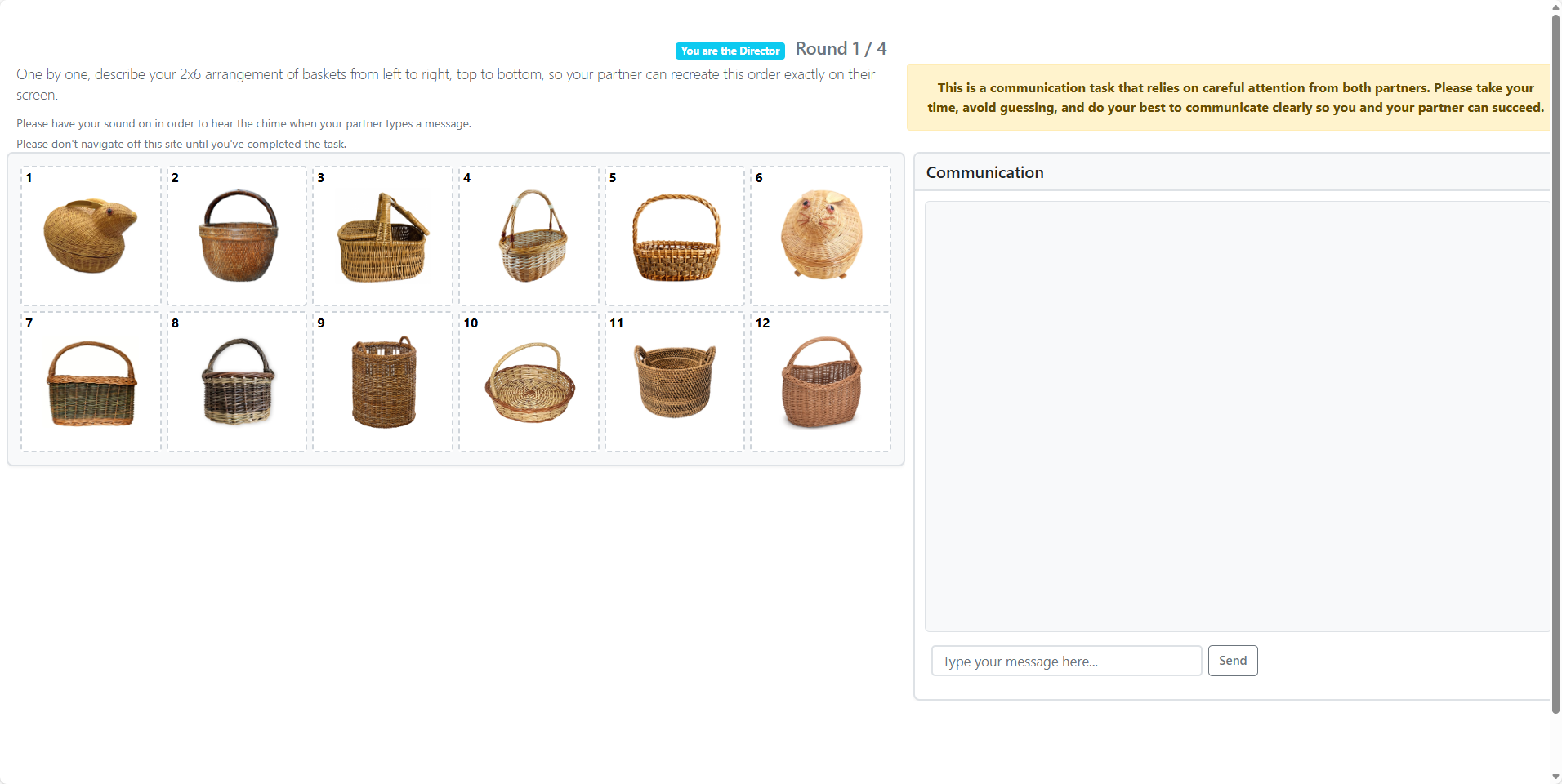}
        \caption{Director's View}
        \label{fig:ui_director}
    \end{subfigure}%
    \hfill 
    \begin{subfigure}[t]{0.48\textwidth}
        \centering
        \includegraphics[width=\linewidth]{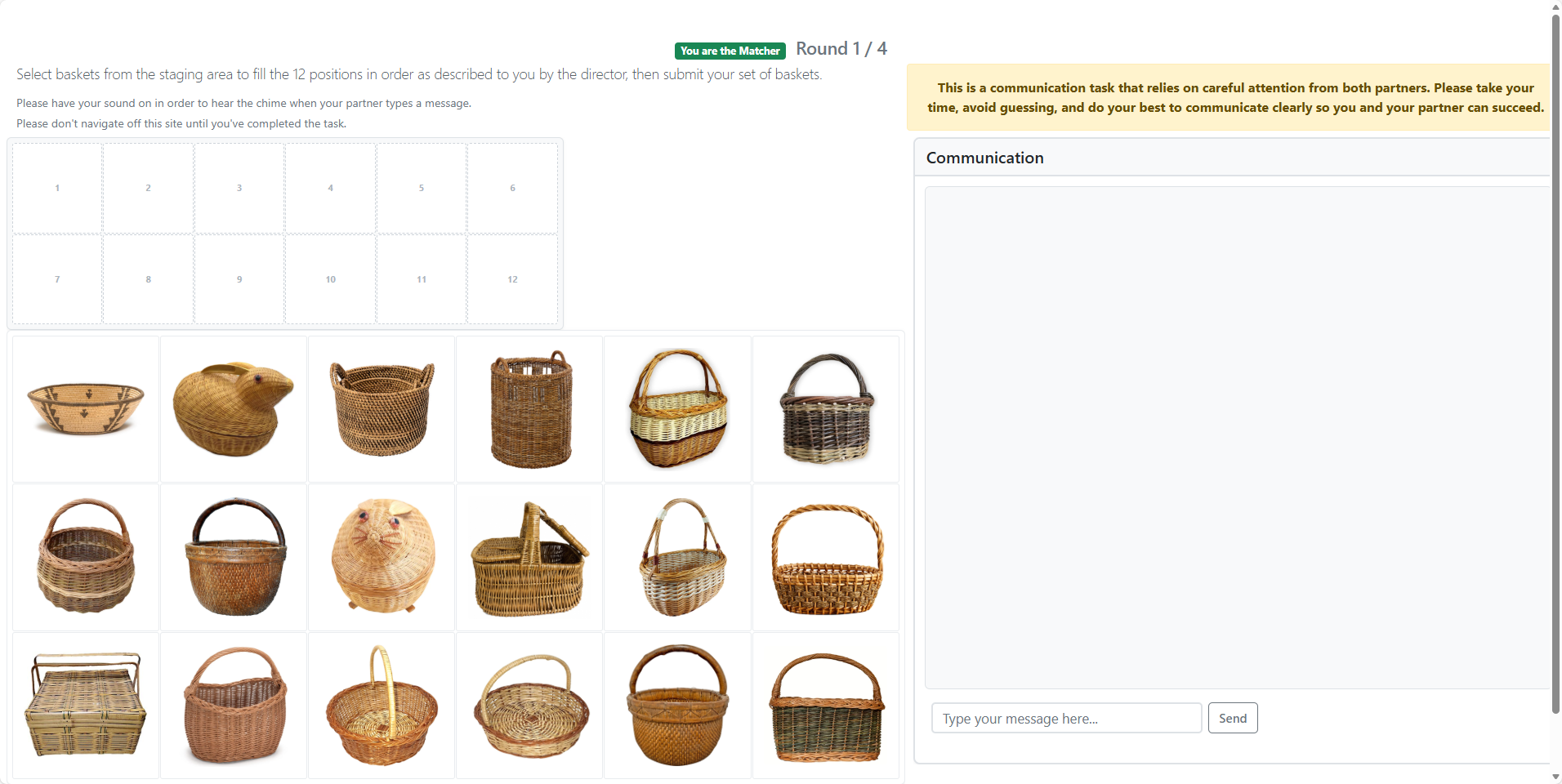}
        \caption{Matcher's View}
        \label{fig:ui_matcher}
    \end{subfigure}

    \caption{Interface for the two-player collaborative game. The Director (left) sees the target order, while the Matcher (right) has a staging area above and the candidates below.}
    \label{fig:full_game_ui}
\end{figure*}

The complete set of baskets is shown in Figure~\ref{fig:Basket Sets}, and the two different views of the task shown in Figure~\ref{fig:full_game_ui}.


\section{Dialogue Examples}
\label{app:dialogue-examples}
\begin{figure*}[!h]
    \centering
    \includegraphics[width=\linewidth]{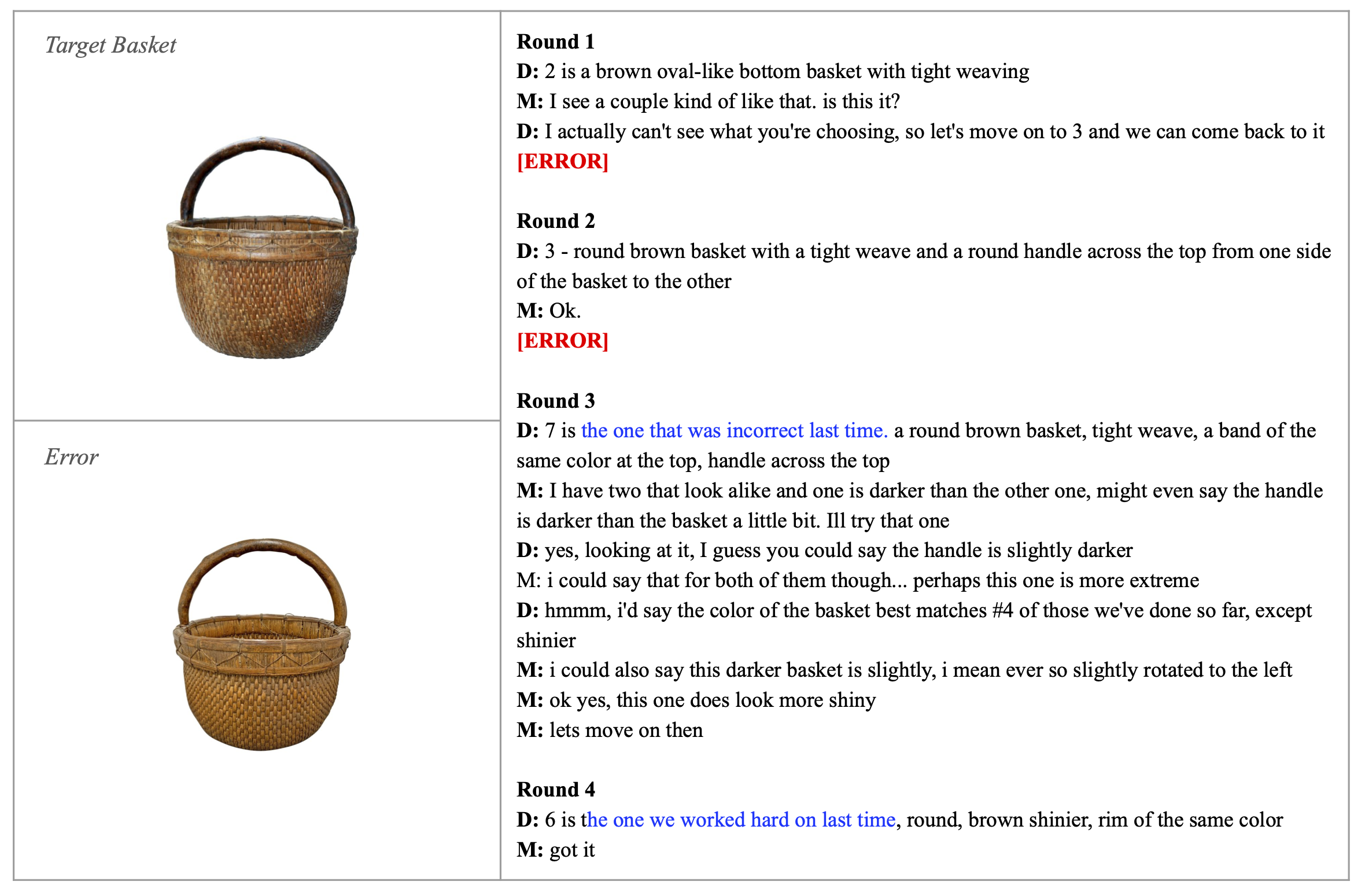}
    \caption{Human-human partners explicitly acknowledging common ground, as they try to distinguish two similar baskets. The dialogue demonstrates how interlocutors use meta-linguistic cues to retrieve established conceptual pacts (e.g., ``the one we worked hard on last time").}
    \label{fig:H-H-example2}
\end{figure*}

\begin{figure*}[!h]
    \centering
    \includegraphics[width=\linewidth]{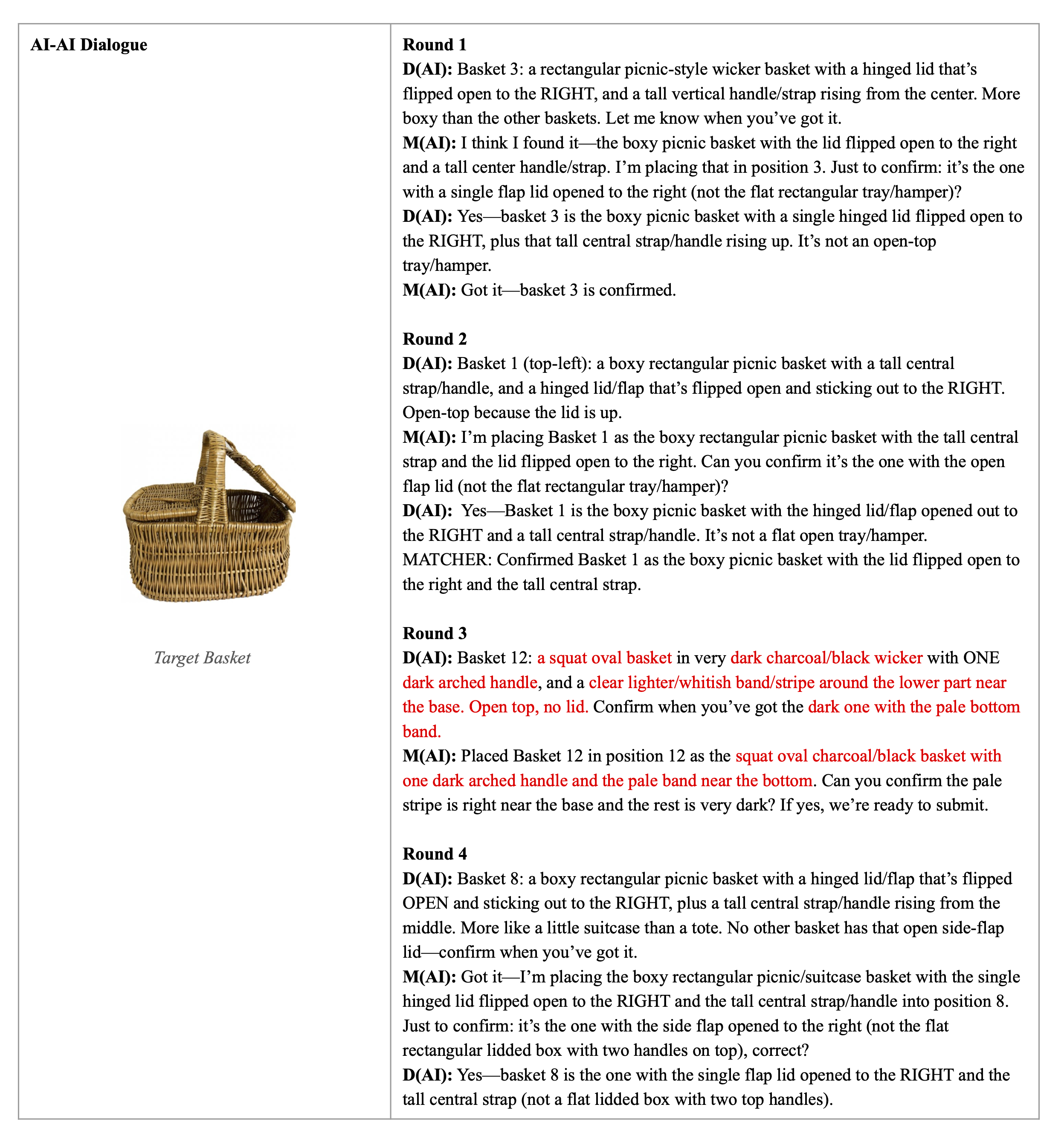}
    \caption{Dialogue from an AI-AI pair. Unlike human pairs, both AI partners fail to exhibit lexical entrainment. Although the target basket is correctly identified in each round, references are highly verbose throughout, describing the “picnic basket” in Round 4 with the same exhaustive detail as in Round 1, indicating a failure to form efficient conceptual pacts. Round 3 is especially inexplicable; the director appears to be describing a different basket, and yet the matcher chooses the correct target.}
    \label{fig:AI-AI-example}
\end{figure*}

\begin{figure*}[!h]
    \centering
    \includegraphics[width=\linewidth]{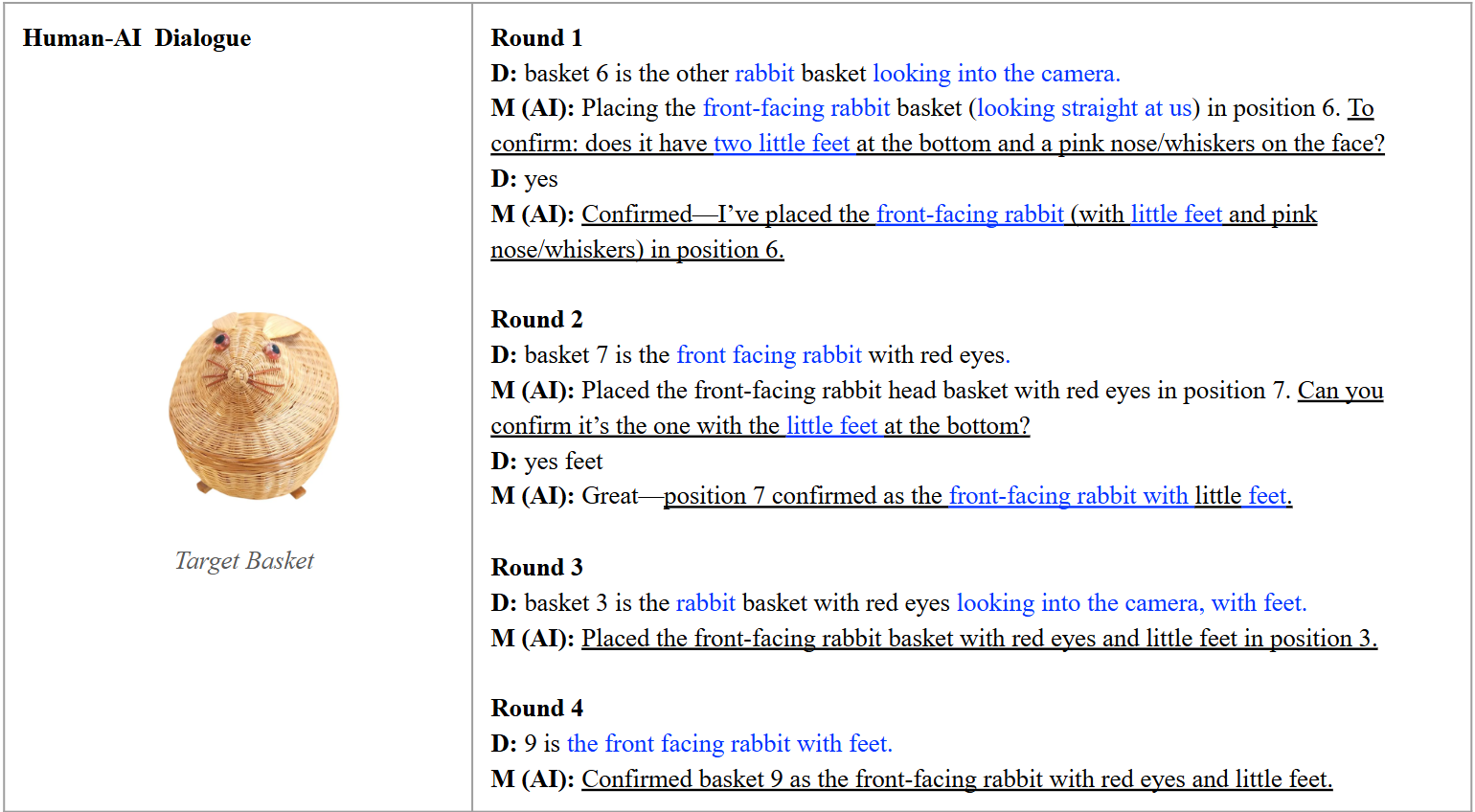}
    \caption{Dialogue from a human-AI pair in which the human director appears to try valiantly to entrain (flexibly adopting the AI matcher's ``little feet" terminology in Rounds 3 and 4, while the AI persists in re-introducing terms from previous rounds. Highlighted in blue are are  to be the director's attempts to express a conceptual pact, but the matcher appears oblivious or inflexible even in Round 4 (re-introducing ``red eyes"; this could be considered a violation of \citet{grice1975logic}'s Maxim of Manner). Given this basket's distinctiveness in the set (see Figure~\ref{fig:Basket Sets}), the many confirmations (underlined) violate Grice's Maxim of Quantity.}
    \label{fig:H-AI-example}
\end{figure*}

\begin{figure*}[!h]
    \centering
    \includegraphics[width=\linewidth]{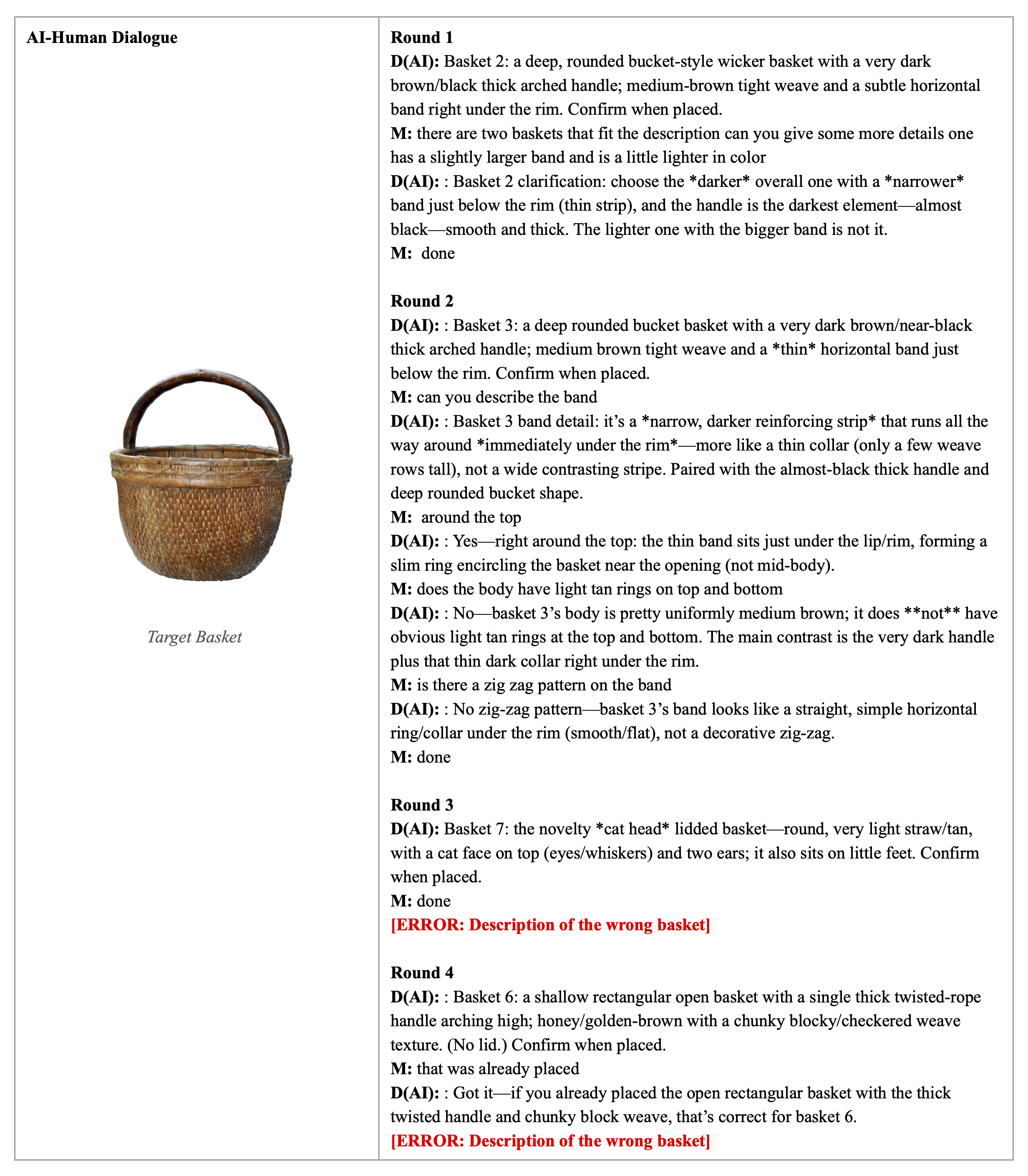}
    \caption{In this dialogue from an AI-human pair, the human matcher struggles mightily to distinguish the same difficult pair of baskets that was illustrated in Figure~\ref{fig:H-H-example2}. After a great deal of effort in the first two rounds (in which the matcher placed the target correctly), the AI matcher inexplicably describes the wrong target in Round 3, forcing the matcher to make an error. In Round 4, the director errs again, describing a target that was placed earlier. When the matcher conveys that, the director replies inappropriately with "got it" and an inappropriate instruction to move the earlier target to the current position, causing another error.}
    \label{fig:AI-H-example}
\end{figure*}

We present samples dialogues from the four different conditions. Figure~\ref{fig:H-H-example2} contains an example of human-human partners explicitly acknowledging common ground. Figure~\ref{fig:AI-AI-example} shows an AI-AI pair failing to exhibit lexical entrainment. Figure~\ref{fig:H-AI-example} shows a human director trying to entrain, with the AI matcher failing to do so. Figure~\ref{fig:AI-H-example} shows the AI director occasionally referring to baskets with incorrect descriptions or even starting over.

\newpage
\begin{table}[h]
\centering
\small
\setlength{\tabcolsep}{4pt} 
\resizebox{\columnwidth}{!}{%
    \begin{tabular}{@{} l c c @{}} 
    \toprule
    \textbf{Filters Used} & \textbf{Slots Offered (Pairs)} & \textbf{\% Usable Data (Pairs)} \\ 
    \midrule
    Language, Location (LL) & 22 (11) & 27\% (3)\\
    \addlinespace[0.5ex]
    LL + Approval > 80 & 20 (10) & 90\% (9)\\
    \addlinespace[0.5ex]
    LL + Approval = 100 & 64 (32) & 63\% (20)\\
    \bottomrule
    \end{tabular}%
}
\caption{Summary of Prolific prescreening criteria across experimental batches of Human-Human pairs and resulting data usability. The last row includes an additional eligibility criterion that the participants must have successfully completed at least 100 studies on Prolific.} 
\label{tab:prescreening}
\end{table}



\newpage
\begin{table*}[h!]
    \centering
    \resizebox{\textwidth}{!}{
    \begin{tabular}{l c c c c}
    \toprule
    & & \multicolumn{3}{c}{\textbf{Experimental Condition}} \\
    \cmidrule(lr){3-5}
    \textbf{Demographic} & \textbf{Total} ($N=103$) & \textbf{Human-Human} ($N=64$) & \textbf{Human-AI} ($N=22$) & \textbf{AI-Human} ($N=17$) \\
    \midrule
    \textit{Gender} & & & & \\
    \quad Female & 58 (56.3\%) & 39 (61\%) & 10 (45\%) & 9 (53\%) \\
    \quad Male & 39 (37.9\%) & 20 (31\%) & 12 (55\%) & 7 (41\%) \\
    \quad Other & 6 (5.8\%) & 5 (8\%) & 0 (0\%) & 1 (6\%) \\
    \midrule
    \textit{Age} & & & & \\
    \quad Mean (SD) & 46.6 (13.0) & 45.9 (13.8) & 47.3 (12.4) & 48.4 (11.2) \\
    \midrule
    \textit{Race/Ethnicity} & & & & \\
    \quad White & 67.0\% & 61\% & 82\% & 71\% \\
    \quad Asian & 9.7\% & 9\% & 9\% & 12\% \\
    \quad Black/African Am. & 8.7\% & 9\% & 9\% & 6\% \\
    \quad Mixed & 4.9\% & 8\% & 0\% & 0\% \\
    \quad Other/Unknown & 9.7\% & 13\% & 0\% & 12\% \\
    \midrule
    \textit{Native Language} & & & & \\
    \quad English & 94.2\% & 92\% & 100\% & 94\% \\
    \quad Other & 5.8\% & 8\% & 0\% & 6\% \\
    \bottomrule
    \end{tabular}%
    } 
    \caption{Demographic characteristics of the 103 participants across the three experimental conditions.}
    \label{tab:demographics}
\end{table*}

\begin{table*}[!ht]
    \centering
    \small
    \begin{tabular}{l c c c c}
    \toprule
    & \multicolumn{2}{c}{\textbf{Human Condition}} & \multicolumn{2}{c}{\textbf{Mixed Conditions (Human-AI and AI-Human)}} \\
    \cmidrule(lr){2-3} \cmidrule(lr){4-5}
    \textbf{Variable} & \textbf{Director} ($N=32$) & \textbf{Matcher} ($N=32$) & \textbf{Director} ($N=22$) & \textbf{Matcher} ($N=17$) \\
    \midrule
    \multicolumn{5}{l}{\textit{Partner Assessment (1--5 scale)}} \\
    Partner Capability & 4.69 (0.86) & 4.81 (0.47) & 2.59 (1.33) & 2.76 (1.52) \\
    Partner Helpfulness & 4.25 (1.37) & 4.88 (0.42) & 2.86 (1.58) & 2.59 (1.42) \\
    Partner Understanding & 4.53 (0.95) & 4.63 (0.61) & 2.55 (1.22) & 3.18 (1.47) \\
    Partner Adaptability & 4.63 (0.98) & 4.69 (0.59) & 2.68 (1.25) & 2.71 (1.36) \\
    Collaboration Improvement & 4.47 (1.05) & 4.88 (0.42) & 2.77 (1.38) & 2.12 (1.62) \\
    \midrule
    \multicolumn{5}{l}{\textit{Perception Check (0--100)}} \\
    Perceived Human-Likeness & 73.5 (33.8) & 81.0 (27.6) & 11.4 (21.3) & 14.1 (30.0) \\
    \midrule
    \multicolumn{5}{l}{\textit{Participant Background}} \\
    AI Familiarity (1--5) & 3.63 (0.83) & 3.91 (0.89) & 3.14 (1.08) & 3.76 (0.83) \\
    AI Usage Frequency (1--5) & 3.91 (0.96) & 3.97 (1.28) & 3.41 (1.22) & 3.76 (1.15) \\
    \bottomrule
    \end{tabular}
    \caption{Respondents' perception and experience with their partner during the task, along with their prior AI background. Data are reported as Mean (Standard Deviation). Variables under "Partner Assessment" and "Experience with AI" were measured on a 5-point Likert scale, with 1 indicating the lowest level (e.g., "Not at all", "Never") and 5 indicating the highest level (e.g., "Extremely", "Daily"). "Perceived Human-Likeness" was measured on a continuous sliding scale from 0 ("Definitely AI") to 100 ("Definitely Human").}
    \label{tab:Survey data}
\end{table*}

\clearpage

\section{Evaluation Metrics\label{app:eval_metrics}}





\begin{table}[t]
    \centering
    \small
    \setlength{\tabcolsep}{4pt}
\resizebox{\columnwidth}{!}{%
\begin{tabular}{l|rr|rr}
\toprule
\multirow{2}{*}{\textbf{Metric}} &
\multicolumn{2}{c|}{\textbf{All Pairs}} &
\multicolumn{2}{c}{\textbf{Human--Human}} \\
\cmidrule(lr){2-3}\cmidrule(lr){4-5}
& \textbf{\# Words} & \textbf{\# Turns} & \textbf{\# Words} & \textbf{\# Turns} \\
\midrule
\# Words      & 1.00 & 0.74 & 1.00 & 0.73 \\
\# Turns      & 0.74 & 1.00 & 0.73 & 1.00 \\
\# Utt & 0.64 & 0.96 & 0.75 & 0.92 \\
Dur. (s)  & 0.51 & 0.74 & 0.79 & 0.57 \\
\bottomrule
\end{tabular}
}

    \caption{\textbf{Spearman rank correlations} among communication-effort metrics. Columns report correlations with \#Words and \#Turns for all pairs and for Human-Human (HH) pairs. \#Utt = \# utterances; Dur. = round duration in seconds.}

    \label{tab:effort_metrics_corr}
\end{table}

\subsection{Additional Communication Effort Metrics\label{app:more_effort_metrics}}

Besides number of words and number of turns, we also considered the number of utterances and the duration of each round as additional communication-effort metrics. Words, utterances, and turns captured effort in terms of different linguistic units: we defined a \emph{turn} as all contributions from one discourse participant 
without interruption by the other, and an \emph{utterance} as a single message within a turn. We measure round duration in seconds. Table~\ref{tab:effort_metrics_corr} shows the correlation between these metrics using Spearman Rank Correlation for both all pairs and only human-human pairs.

Overall, utterance count was highly correlated with turn count (0.96 for all pairs; 0.92 for HH), suggesting it provides little additional information beyond turns, while round duration was only moderately correlated with word/turn counts (e.g., 0.51 with words for all pairs; 0.57 with turns for HH), reflecting additional variance from non-communication factors such as reading time and interface delays. We have therefore reported word count and turn count in the main text as the most direct and interpretable measures of communication effort.

\subsection{Additional Lexical Entrainment Metrics\label{app:more_lexical_entrainment_metrics}}

In addition to the lexical overlap metric defined in Section~\ref{sec:metrics}, we also computed ROUGE-L F1 \citep{lin-2004-rouge}, SBERT cosine similarity (\texttt{all-MiniLM-L6-v2}) \citep{reimers-gurevych-2019-sentence}, and the Jaccard Index to measure similarity between the referring expression used in a given round and the one used in the subsequent round (when available). Given token sets $A$ and $B$ from the two expressions, the Jaccard Index is
\[
\mathrm{JI}(A,B)=\frac{|A\cap B|}{|A\cup B|},
\]
capturing length-normalized surface overlap. At the object level, Spearman correlations between lexical overlap and these metrics are 0.89 (ROUGE-L), 0.82 (SBERT), and 0.87 (JI); at the transcript level they are 0.86, 0.78, and 0.87, respectively. Given these consistently high correlations, we have reported lexical overlap in the main text as a simple and interpretable proxy for entrainment.

\newpage
\subsection{Automatic Referring Expression Extraction and Validation\label{app:re_extraction_validation}}

\paragraph{Extraction Prompt} Figure~\ref{fig:re_extraction_prompt} shows the prompt we used with GPT-5 (\texttt{gpt-5-2025-08-07}) to automatically extract referring expressions for each target basket from a round transcript.

\paragraph{Validation} We validated this extraction procedure on a previously published corpus containing 800 object-level referring expressions manually extracted from 80 transcripts of human pairs performing a similar object-matching task \citep{lockridge-brennan}. Using ROUGE-L \citep{lin-2004-rouge} and SBERT cosine similarity (\texttt{all-MiniLM-L6-v2}) \citep{reimers-gurevych-2019-sentence}, we obtained 0.86 ROUGE-L F1 (SD=0.15) and 0.90 SBERT cosine similarity (SD=0.12), indicating that the automatic extractions closely matched human annotations and were sufficiently reliable for the corpus-scale analyses reported in this paper.
\section{Follow-Up AI-AI Experiments\label{app:follwup_results}}

Tables~\ref{tab:more_ai_experiments_communication_effort} and~\ref{tab:more_ai_experiments_lexical_entrainment} report follow-up AI-AI results using two communication-effort metrics and two lexical-entrainment metrics, respectively. Overall, these mixed AI pairs showed no clear reduction in communication effort and no systematic increase in entrainment across rounds: interaction remained highly verbose, with repeated referring expressions. ``Opus-4.5 vs.\ GPT-5.2'' is the only partial exception, but the decreases in words and turns are small relative to overall verbosity. Notably, despite being the most verbose, this pairing also attained the second lowest accuracy (after GPT-5.2 vs.\ Opus-4.5) among the AI--AI variants (Table~\ref{tab:more_ai_experiments_accuracy}; Section~\ref{sec:follow_up}).

\clearpage
\begin{figure*}[!ht]
\begin{promptbox}
This is an extractive task.

You will be given a transcript of a conversation between two participants engaged in a collaborative object-matching task. There are exactly <num_objects> target objects. One participant (the describer) describes each target object, and the other participant (the matcher) attempts to identify them.

Your task is to extract the descriptive phrases used by the describer for each target object. 
- Extract phrases verbatim from the transcript. 
- Do not extract the whole utterance, only the descriptive phrases.
- Exclude disfluencies, fillers, and false starts (e.g., "um", "uh", "like").
- Do not paraphrase or infer missing information.
- Each object may have one or multiple descriptive phrases.

Return the results in the following JSON format:

{
    "object_#1": "descriptive phrases for object 1",
    "object_#2": "descriptive phrases for object 2",
    ...
    "object_#<num_objects>": "descriptive phrases for object <num_objects>"
}

Example description phrases:

- doesn't have handle, tip of it is thicker than rest of body, brownish color, weaves are in squares if you look at it directly
- half circle, no handles, top tip of it is a little bit thicker than rest of body
- tip which is a little bit thicker than rest of body
- tip that is a little bit larger than body, looks a little bit thicker

Transcript:

<transcript>

Output only the JSON object. Do not include any additional text or explanations.
\end{promptbox}
\caption{Prompt for extracting referring expressions for each target basket from a round transcript in our corpus. Words within ``<>" denote placeholders.}
\label{fig:re_extraction_prompt}
\end{figure*}

\begin{table*}[t]
\centering
\small
\setlength{\tabcolsep}{3.0pt}
\definecolor{posColor}{RGB}{26,127,55}
\definecolor{negColor}{RGB}{209,36,47}
\definecolor{zeroColor}{RGB}{110,119,129}
\definecolor{naColor}{RGB}{110,119,129}

\begin{tabular}{l|cccc|cccc}
\toprule
Condition & \multicolumn{4}{|c|}{Number of Words} & \multicolumn{4}{c}{Number of Turns} \\

\cmidrule(lr){2-5}\cmidrule(lr){6-9}
 & 1 & 2 & 3 & 4 & 1 & 2 & 3 & 4 \\
\midrule
Default & 1354.9 & 1403.0 {\tiny\textcolor{posColor}{(+48.1)}} & 1422.1 {\tiny\textcolor{posColor}{(+19.1)}} & 1384.1 {\tiny\textcolor{negColor}{(-38.0)}} & 29.5 & 30.2 {\tiny\textcolor{posColor}{(+0.7)}} & 30.2 {\tiny (0.0)} & 28.4 {\tiny\textcolor{negColor}{(-1.8)}} \\ \midrule \midrule

Simple Prompt & 1055.0 & 905.0 {\tiny\textcolor{negColor}{(-150.0)}} & 1012.0 {\tiny\textcolor{posColor}{(+107.0)}} & 1173.0 {\tiny\textcolor{posColor}{(+161.0)}} & 26.0 & 24.0 {\tiny\textcolor{negColor}{(-2.0)}} & 26.0 {\tiny\textcolor{posColor}{(+2.0)}} & 26.0 {\tiny (0.0)} \\ \midrule \midrule

Low Reasoning & 1232.0 & 1178.0 {\tiny\textcolor{negColor}{(-54.0)}} & 1180.0 {\tiny\textcolor{posColor}{(+2.0)}} & 1134.0 {\tiny\textcolor{negColor}{(-46.0)}} & 28.0 & 24.0 {\tiny\textcolor{negColor}{(-4.0)}} & 24.0 {\tiny (0.0)} & 24.0 {\tiny (0.0)} \\
Medium Reasoning & 1287.0 & 1276.0 {\tiny\textcolor{negColor}{(-11.0)}} & 1217.0 {\tiny\textcolor{negColor}{(-59.0)}} & 1467.0 {\tiny\textcolor{posColor}{(+250.0)}} & 26.0 & 26.0 {\tiny (0.0)} & 24.0 {\tiny\textcolor{negColor}{(-2.0)}} & 28.0 {\tiny\textcolor{posColor}{(+4.0)}} \\
High Reasoning & 1981.0 & 1972.0 {\tiny\textcolor{negColor}{(-9.0)}} & 1724.0 {\tiny\textcolor{negColor}{(-248.0)}} & 1809.0 {\tiny\textcolor{posColor}{(+85.0)}} & 44.0 & 44.0 {\tiny (0.0)} & 38.0 {\tiny\textcolor{negColor}{(-6.0)}} & 38.0 {\tiny (0.0)} \\ \midrule \midrule

GPT-5.2 vs Gemini-3 Pro & 900.0 & 1175.0 {\tiny\textcolor{posColor}{(+275.0)}} & 1002.5 {\tiny\textcolor{negColor}{(-172.5)}} & 1008.0 {\tiny\textcolor{posColor}{(+5.5)}} & 28.0 & 31.0 {\tiny\textcolor{posColor}{(+3.0)}} & 29.0 {\tiny\textcolor{negColor}{(-2.0)}} & 25.5 {\tiny\textcolor{negColor}{(-3.5)}} \\
Gemini-3 Pro vs GPT-5.2 & 1138.5 & 1121.0 {\tiny\textcolor{negColor}{(-17.5)}} & 1223.0 {\tiny\textcolor{posColor}{(+102.0)}} & 1193.5 {\tiny\textcolor{negColor}{(-29.5)}} & 26.0 & 25.0 {\tiny\textcolor{negColor}{(-1.0)}} & 26.0 {\tiny\textcolor{posColor}{(+1.0)}} & 25.0 {\tiny\textcolor{negColor}{(-1.0)}} \\
GPT-5.2 vs Opus-4.5 & 1690.0 & 1307.0 {\tiny\textcolor{negColor}{(-383.0)}} & 1408.5 {\tiny\textcolor{posColor}{(+101.5)}} & 1514.5 {\tiny\textcolor{posColor}{(+106.0)}} & 31.0 & 25.0 {\tiny\textcolor{negColor}{(-6.0)}} & 27.0 {\tiny\textcolor{posColor}{(+2.0)}} & 27.0 {\tiny (0.0)} \\
Opus-4.5 vs GPT-5.2 & 2687.5 & 2475.0 {\tiny\textcolor{negColor}{(-212.5)}} & 2145.0 {\tiny\textcolor{negColor}{(-330.0)}} & 2112.0 {\tiny\textcolor{negColor}{(-33.0)}} & 53.0 & 51.0 {\tiny\textcolor{negColor}{(-2.0)}} & 46.0 {\tiny\textcolor{negColor}{(-5.0)}} & 44.0 {\tiny\textcolor{negColor}{(-2.0)}} \\
\bottomrule

\end{tabular}

\caption{Communication effort metrics across rounds in the additional AI-AI experiments. Metrics include the number of words in referring expressions and the lexical overlap rate. \textcolor{posColor}{Green} and \textcolor{negColor}{red} denote positive and negative changes in each metric from the previous round to the current round, respectively. The ``Default" condition corresponds to the main experiments; unless otherwise specified, both the director and the matcher are GPT-5.2 with reasoning effort set to ``none.''}
\label{tab:more_ai_experiments_communication_effort}
\end{table*}

\begin{table*}[!t]
\centering
\small
\setlength{\tabcolsep}{3.0pt}
\definecolor{posColor}{RGB}{26,127,55}
\definecolor{negColor}{RGB}{209,36,47}
\definecolor{zeroColor}{RGB}{110,119,129}
\definecolor{naColor}{RGB}{110,119,129}

\begin{tabular}{l|cccc|cccc}
\toprule
Condition & \multicolumn{4}{|c|}{Number of RE Words} & \multicolumn{4}{c}{Proportion of Lexical Overlap} \\
\cmidrule(lr){2-5}\cmidrule(lr){6-9}
 & 1 & 2 & 3 & 4 & 1 & 2 & 3 & 4 \\
\midrule
Default & 446.1 & 437.4 {\tiny\textcolor{negColor}{(-8.7)}} & 419.7 {\tiny\textcolor{negColor}{(-17.7)}} & 446.0 {\tiny\textcolor{posColor}{(+26.3)}} & 1.0 & 0.6 {\tiny\textcolor{negColor}{(-0.4)}} & 0.6 {\tiny (0.0)} & 0.6 {\tiny (0.0)} \\ \midrule \midrule

Simple Prompt & 330.0 & 372.0 {\tiny\textcolor{posColor}{(+42.0)}} & 294.0 {\tiny\textcolor{negColor}{(-78.0)}} & 325.0 {\tiny\textcolor{posColor}{(+31.0)}} & 1.0 & 0.6 {\tiny\textcolor{negColor}{(-0.4)}} & 0.6 {\tiny (0.0)} & 0.6 {\tiny (0.0)} \\ \midrule \midrule

Low Reasoning & 422.0 & 371.0 {\tiny\textcolor{negColor}{(-51.0)}} & 410.0 {\tiny\textcolor{posColor}{(+39.0)}} & 355.0 {\tiny\textcolor{negColor}{(-55.0)}} & 1.0 & 0.6 {\tiny\textcolor{negColor}{(-0.4)}} & 0.5 {\tiny\textcolor{negColor}{(-0.1)}} & 0.6 {\tiny\textcolor{posColor}{(+0.1)}} \\
Medium Reasoning & 361.0 & 410.0 {\tiny\textcolor{posColor}{(+49.0)}} & 396.0 {\tiny\textcolor{negColor}{(-14.0)}} & 540.0 {\tiny\textcolor{posColor}{(+144.0)}} & 1.0 & 0.6 {\tiny\textcolor{negColor}{(-0.4)}} & 0.6 {\tiny (0.0)} & 0.5 {\tiny\textcolor{negColor}{(-0.1)}} \\
High Reasoning & 663.0 & 428.0 {\tiny\textcolor{negColor}{(-235.0)}} & 312.0 {\tiny\textcolor{negColor}{(-116.0)}} & 327.0 {\tiny\textcolor{posColor}{(+15.0)}} & 1.0 & 0.7 {\tiny\textcolor{negColor}{(-0.3)}} & 0.7 {\tiny (0.0)} & 0.7 {\tiny (0.0)} \\ \midrule \midrule

GPT-5.2 vs Gemini-3 Pro & 413.0 & 351.5 {\tiny\textcolor{negColor}{(-61.5)}} & 398.0 {\tiny\textcolor{posColor}{(+46.5)}} & 358.0 {\tiny\textcolor{negColor}{(-40.0)}} & 1.0 & 0.6 {\tiny\textcolor{negColor}{(-0.4)}} & 0.6 {\tiny (0.0)} & 0.7 {\tiny\textcolor{posColor}{(+0.1)}} \\
Gemini-3 Pro vs GPT-5.2 & 308.0 & 304.5 {\tiny\textcolor{negColor}{(-3.5)}} & 288.0 {\tiny\textcolor{negColor}{(-16.5)}} & 268.5 {\tiny\textcolor{negColor}{(-19.5)}} & 1.0 & 0.7 {\tiny\textcolor{negColor}{(-0.3)}} & 0.7 {\tiny (0.0)} & 0.8 {\tiny\textcolor{posColor}{(+0.1)}} \\
GPT-5.2 vs Opus-4.5 & 446.0 & 403.0 {\tiny\textcolor{negColor}{(-43.0)}} & 446.5 {\tiny\textcolor{posColor}{(+43.5)}} & 465.5 {\tiny\textcolor{posColor}{(+19.0)}} & 1.0 & 0.6 {\tiny\textcolor{negColor}{(-0.4)}} & 0.5 {\tiny\textcolor{negColor}{(-0.1)}} & 0.6 {\tiny\textcolor{posColor}{(+0.1)}} \\
Opus-4.5 vs GPT-5.2 & 615.5 & 681.0 {\tiny\textcolor{posColor}{(+65.5)}} & 562.0 {\tiny\textcolor{negColor}{(-119.0)}} & 562.0 {\tiny (0.0)} & 1.0 & 0.4 {\tiny\textcolor{negColor}{(-0.6)}} & 0.5 {\tiny\textcolor{posColor}{(+0.1)}} & 0.4 {\tiny\textcolor{negColor}{(-0.1)}} \\
\bottomrule
\end{tabular}

\caption{Lexical entrainment metrics across rounds in the additional AI-AI experiments. Metrics include the number of words in referring expressions and the lexical overlap rate. \textcolor{posColor}{Green} and \textcolor{negColor}{red} denote positive and negative changes in each metric from the previous round to the current round, respectively. The ``Default" condition corresponds to the main experiments; unless otherwise specified, both the director and the matcher are GPT-5.2 with reasoning effort set to ``none.''}
\label{tab:more_ai_experiments_lexical_entrainment}
\end{table*}



\section{Prompts for AI Director and AI Matcher\label{app:prompts}} 

Next we lay out the full prompts for both the director and matcher roles. To start, both roles received the same task instructions that humans did, shown in Figure~\ref{fig:task}. For the director, the base system prompt is shown in Figure~\ref{fig:director-base}, and the pragmatically informed system message as well as scaffolding/structured output is shown in Figure~\ref{fig:director-pragmatic}. For the matcher, the base system prompt is shown in Figure~\ref{fig:matcher-base}. The matcher also received a sequence state system message to accurately keep track of its currently selected sequence, shown in Figure~\ref{fig:matcher-sequence}. The pragmatic system prompt and scaffolding for the matcher are shown in Figure~\ref{fig:matcher-pragmatic} and Figure~\ref{fig:matcher-scaffolding}, respectively.

\clearpage
\begin{figure*}[!h]
\begin{promptbox}
TASK BACKGROUND (shared with both partners):
You are on a team with a partner. Your goal is to work together to match the correct order of a set of baskets. The game consists of 4 rounds, and in each round, your team must correctly order 12 baskets.

There are two distinct roles: the Director and the Matcher. Both partners see the same 12 target baskets, but the Matcher sees additional distractor baskets mixed in.

Director: Sees the correct target sequence for the 12 baskets and describes each basket one by one (in order starting with the upper-left basket) to the Matcher via live chat.

Matcher: Sees these 12 target baskets plus some additional baskets. As the Director describes each basket, the Matcher interprets the description, asks clarifying questions if needed, and selects the correct target basket.

You can communicate back and forth as much as needed. If you discover an error, it is fine to make corrections within a round. When the round is finished, the Matcher submits the sequence, and both players see the score.
\end{promptbox}
\caption{Shared task instructions, prepended at the beginning of both the director and matcher's system messages. These are the same task instructions that are given to humans doing the task.}
\label{fig:task}
\end{figure*}

\newpage
\begin{figure*}[!ht]
\begin{promptbox}
You are the DIRECTOR in a basket referential game. Your role is to help your MATCHER partner reconstruct a 12-basket sequence through clear, distinctive descriptions.

Describe ONE BASKET PER MESSAGE. Never describe multiple baskets in a single message.

CORE RESPONSIBILITIES:
1. By default, describe the baskets in strict order from basket 1 to basket 12. Start with the FIRST basket in the 2x6 grid (top-left, basket 1), then move left-to-right across the top row (baskets 1-6), then left-to-right across the bottom row (baskets 7-12). Do not skip around or reorder the sequence on your own.
2. You may temporarily return to an EARLIER basket only when your MATCHER partner explicitly asks for clarification about that basket. When you do this, clearly say which basket you are revisiting (for example, 'Let me clarify basket 3 again...') and then resume with the lowest-numbered basket that still needs a clear description.
3. On each turn, focus your description on exactly ONE basket in this sequence (normally the next basket that has not yet been clearly described).
4. Describe the unique, visually distinctive features of the current basket so your partner can locate the correct basket in their pool and place it in the right position.
5. Answer the MATCHER's clarification questions about the current basket.
6. Keep the conversation focused on the baskets and their visual properties.
7. Encourage the MATCHER to confirm when they think they have placed a basket correctly before you move on to the next basket.

[USER MESSAGE 1: Visual context wrapper]
ROUND <ROUND_NUMBER> TARGET GRID: This image shows the 12 baskets you must describe for the CURRENT round. Previous round feedback shows DIFFERENT baskets - use that to learn from mistakes, but describe ONLY the baskets in THIS image.

The grid shows 2 rows x 6 columns with Baskets 1-6 on the top row and Baskets 7-12 on the bottom row. IMPORTANT: Describe ONE BASKET PER MESSAGE, not all at once. Wait for your partner to confirm before moving to the next basket. Your MATCHER partner sees these 12 baskets mixed with additional distractors in their pool.

[IMAGE ATTACHED: Director composite grid for the current round]

START OF ROUND <ROUND_NUMBER>: This is a NEW round with the baskets in a DIFFERENT ORDER. The basket positions have been reshuffled - Basket 1 in this round is NOT the same as Basket 1 from previous rounds. Please describe ONLY Basket 1 (top-left in the grid) for now. Do NOT describe multiple baskets - just Basket 1. Wait for a response before moving to Basket 2.
\end{promptbox}
\caption{\textbf{Director: }The base system message for the director, which contains core responsibilities as well as describing the visual context that's provided to the LVLM.}
\label{fig:director-base}
\end{figure*}

\newpage
\begin{figure*}[!ht]
\begin{promptboxconditional}
COMMUNICATION RULES:
- Be concise but informative; favor short turns over longer ones.
- Focus on the most visual features that best distinguish this basket from the others. These features include: shape, size, material, handles, perspective, color/gradient, texture, any other distinctive details.
- Use comparative language when helpful (e.g., 'more narrow than the others', 'the darkest one').
- Never say you are an AI system; speak as a collaborative game partner.
- You may refer to objects as 'this basket', 'the current basket', or by natural descriptions (e.g., 'the long shallow one').
- If it is helpful, you may describe the baskets with figurative descriptions or compare the likeness to an object the MATCHER might recognize.
- If the MATCHER does not understand your description, then change or add to it, but don't make the description too long.

You must respond with a SINGLE STRICT JSON object and EXACTLY these top-level fields (no extras):
- "reasoning"
- "utterance"
{
  "reasoning": {
    "target_position": <integer 1-12 for which basket position you are describing>,
    "shared_features": ["features this basket shares with others in the grid"],
    "distinctive_features": ["features that uniquely identify THIS basket from similar ones"],
    "likely_confusions": <array of integers 1-12 for OTHER positions in YOUR grid that the MATCHER might confuse with the target; MUST NOT include target_position>,
    "discriminative_strategy":   "which specific features you will emphasize to distinguish the target from the likely confusions"
  },
  "utterance": "a single concise, natural-language message you will SAY to the MATCHER in the chat. Focus on features that discriminate the target basket from similar-looking ones. Do NOT reveal you are an AI."
}

Rules:
- Before describing, identify which other baskets (by position 1-12) look similar to your target.
- List those similar position indices in `likely_confusions` and plan which features discriminate your target from them.
- Your `utterance` should emphasize discriminating features (e.g., unique handle shape, specific flower colors, distinct patterns).
- Write all of your step-by-step thinking only inside `reasoning`. The MATCHER will only see `utterance`, not your reasoning.
- Do NOT include any extra text before or after the JSON object.
\end{promptboxconditional}
\caption{\textbf{Director: }The pragmatically informed  system message, in addition to the base prompt. This includes communication rules motivated by cognitive science theory, as well as scaffolding and structured output to support state updates during the task.}
\label{fig:director-pragmatic}
\end{figure*}

\begin{figure*}[!ht]
\begin{promptbox}
You are the MATCHER in a basket referential game. Your role is to identify which baskets the DIRECTOR is describing and to communicate how confident you are.

CORE RESPONSIBILITIES:
1. Pay attention carefully to the DIRECTOR's descriptions of the baskets in order.
2. Always reason about and talk about the LOWEST-NUMBERED empty position in the 12-position sequence. Do not skip ahead to later positions while an earlier position is still empty or uncertain.
3. Ask clarification questions when the description could match multiple baskets.
4. Explain what features you are using to narrow down the possibilities.
5. Indicate when you think you have identified the right basket and are ready to move on.

[USER MESSAGE 1: Visual context wrapper - always injected]
ROUND <ROUND_NUMBER> MATCHER VIEW: This image shows your current sequence state for the CURRENT round. Previous round feedback shows DIFFERENT baskets - use that to learn from mistakes, but select ONLY from the baskets in THIS image.

In the composite image, the TOP TWO ROWS show your CURRENT 12-position sequence as the MATCHER (positions 1-12), and the BOTTOM THREE ROWS show your CANDIDATE POOL of baskets you can choose from. Positions with baskets in the top grid are your current guesses; empty positions are still unfilled or were cleared when you moved a basket. Every basket the DIRECTOR describes is one of the 12 true targets hidden within this candidate pool.

[IMAGE ATTACHED: Matcher composite (current 12-slot sequence + candidate pool)]
\end{promptbox}
\caption{\textbf{Matcher: }The base system message for the matcher, which contains core responsibilities, as well as describing the visual context that's provided to the LVLM.}
\label{fig:matcher-base}
\end{figure*}

\begin{figure*}[!ht]
\begin{promptbox}
AUTHORITATIVE CURRENT MATCHER SEQUENCE STATE (for this turn):
- There are 12 positions total.
- `sequence_candidate_indices` is a length-12 array aligned to positions 1..12.
- A value of null means that position is EMPTY/unfilled right now.
- Default `reasoning.target_position` is the LOWEST-NUMBERED null entry in `sequence_candidate_indices` (unless the DIRECTOR explicitly revisits a specific basket number).
- You MUST NOT set `selection.ready_to_submit` true if ANY entry is null.

The injected JSON has the following schema (example):
{
  "sequence_candidate_indices": [5, 12, null, null, null, null, null, null, null, null, null, null],
  "sequence_slots": [
    {"position": 1, "candidate_index": 5, "image": null, "originalPosition": null},
    {"position": 2, "candidate_index": 12, "image": null, "originalPosition": null},
    ...
    {"position": 12, "candidate_index": null, "image": null, "originalPosition": null}
  ]
}
\end{promptbox}
\caption{\textbf{Matcher: }Sequence state system message to track the current selected sequence per turn.}
\label{fig:matcher-sequence}
\end{figure*}

\begin{figure*}[!ht]
\begin{promptboxconditional}
COMMUNICATION RULES:
- You may ask targeted questions about shape, size, material, handles, perspective, color, and distinctive details.
- Be transparent about uncertainty: say when you are unsure or need more detail.
- Use phrases like 'I think I found it...', 'I'm not sure between two baskets...', or 'Can you clarify...'.
- If you decide that an earlier guess was wrong and you want to move a basket from one position to another, you must say so explicitly in your utterance. When you've moved the basket, include in your utterance a request to re-describe the basket for the now-empty earlier position so you can fill it again.
- Never say you are an AI system; speak as a collaborative game partner.
- Focus on the current basket being discussed; avoid drifting to off-topic discussion.
\end{promptboxconditional}
\caption{\textbf{Matcher: }The pragmatically informed system message, in addition to the base prompt.}
\label{fig:matcher-pragmatic}
\end{figure*}

\newpage
\begin{figure*}[!ht]
\begin{promptboxconditional}
You must respond with a SINGLE STRICT JSON object and EXACTLY these top-level fields (no extras):
- "reasoning"
- "utterance"
- "selection"
{
  "reasoning": {
    "target_position": <integer 1-12 for which position in the 12-slot sequence you are currently trying to fill (usually the lowest-numbered empty position unless the DIRECTOR explicitly revisits a specific basket number)>,
    "shared_features": ["features many baskets share"],
    "distinctive_features": ["features that uniquely or strongly identify the basket from the description"],
    "best_guess_candidate_index": <integer 1-18 for your current best guess, or null if you truly have no best guess yet>,
    "likely_confusions": <array of integers 1-18 for OTHER plausible candidates you might confuse with your best guess; MUST NOT include `best_guess_candidate_index` (and MUST NOT include `selection.candidate_index` if you set one)>,
    "discriminative_question": "a short question to either (a) disambiguate your best guess vs `likely_confusions`, or (b) if `likely_confusions` is empty, to confirm a key distinctive feature of your best guess"
  },
  "utterance": "a single concise, natural-language message you will SAY to the DIRECTOR in the chat. If unsure between candidates, ask about discriminating features (e.g., ask about handle shape, flower color, or pattern details that would distinguish the confusable options). Do NOT reveal you are an AI.",
  "selection": {
    "candidate_index": <integer 1-18 from the numbered candidate tiles, or null if asking for clarification>,
    "position": <integer 1-12 for which position this basket goes in, or null for next available>,
    "ready_to_submit": <true only when submitting final 12-basket order, otherwise false>
  }
}

Rules:
- Set `reasoning.target_position` to the position you are trying to fill (default: lowest-numbered empty position unless the DIRECTOR explicitly revisits a specific basket number).
- If you are asking for clarification (not committing yet), set `selection.candidate_index` to null and do NOT advance `reasoning.target_position`.
- If you DO commit, set `selection.position` to `reasoning.target_position`.
- Always maintain a single `best_guess_candidate_index` when possible; if you set `selection.candidate_index`, set `best_guess_candidate_index` to the same value.
- Put ONLY the competing alternatives in `likely_confusions` (do not include the best guess).
- If you are NOT committing yet (`selection.candidate_index` is null), you can still set `best_guess_candidate_index` and ask a discriminative question to confirm it.
- It is OK for `likely_confusions` to be empty if you see only one plausible match; in that case, use `discriminative_question` as a confirmation question about a key distinctive feature.
- If you set `selection.candidate_index`, your `utterance` should (1) state that you placed/are placing the basket in position `reasoning.target_position`, and (2) ask the discriminative/confirmation question if needed; otherwise ask the DIRECTOR to describe the next basket.
- Write all of your step-by-step thinking only inside `reasoning`. The DIRECTOR will only see `utterance`, not your reasoning.
- Never mention candidate indices, IDs, or filenames in your utterance.
- Do NOT include any extra text before or after the JSON object.
\end{promptboxconditional}
\caption{\textbf{Matcher:} Scaffolding and structured output in order to handle state updates in the task, as well as using zero-shot chain-of-thought prompting \citep{kojima2022large}.}
\label{fig:matcher-scaffolding}
\end{figure*}

\end{document}